\documentclass[10pt,journal,compsoc]{IEEEtran}

\ifCLASSOPTIONcompsoc
  \usepackage[nocompress]{cite}
\else
  \usepackage{cite}
\fi

\usepackage{graphicx}
\usepackage{amssymb}
\usepackage{listings}
\newcommand{\upmark}{$\rightarrow$}
\newcommand{\mean}[1]{$\overline{\mbox{#1}}$}
\newcommand{\textoverline}[1]{$\overline{\mbox{#1}}$}
\makeatletter
\def\blfootnote{\xdef\@thefnmark{}\@footnotetext}
\makeatother

\usepackage{dblfloatfix}    
\usepackage{amsmath}
\usepackage{comment}

\begin{document}
%
\title{Efficient Ladder-style DenseNets \\ for Semantic Segmentation of Large Images}
%
%

\author{Ivan Kre\v{s}o
  \qquad Josip Krapac
  \qquad Sini\v{s}a \v{S}egvi\'{c}\\
  Faculty of Electrical Engineering and Computing\\
  University of Zagreb, Croatia\\
  {\tt\small ivan.kreso@fer.hr \qquad josip.krapac@zalando.de \qquad sinisa.segvic@fer.hr}
}

\markboth{Journal of \LaTeX\ Class Files,~Vol.~14, No.~8, August~2015}%
{Shell \MakeLowercase{\textit{et al.}}: Bare Demo of IEEEtran.cls for Computer Society Journals}
%



\IEEEtitleabstractindextext{

\begin{abstract}
   Recent progress of deep image classification models
   has provided great potential 
   to improve state-of-the-art performance
   in related computer vision tasks.
   However, the transition to semantic segmentation 
   is hampered by strict memory limitations 
   of contemporary GPUs.
   The extent of feature map caching 
   required by convolutional backprop
   poses significant challenges 
   even for moderately sized Pascal images,
   while requiring careful architectural considerations
   when the source resolution is in the megapixel range.
   To address these concerns, we propose a novel
   DenseNet-based ladder-style architecture 
   which features high modelling power 
   and a very lean upsampling datapath.   
   We also propose to substantially reduce
   the extent of feature map caching
   by exploiting inherent spatial efficiency 
   of the DenseNet feature extractor.
   The resulting models deliver high performance 
   with fewer parameters than competitive approaches, 
   and allow training at megapixel resolution 
   on commodity hardware.
   The presented experimental results 
   outperform the state-of-the-art 
   in terms of prediction accuracy and execution speed 
   on Cityscapes, Pascal VOC 2012, 
   CamVid and ROB 2018 datasets.
   Source code will be released upon publication.
\end{abstract}

}

\maketitle

\IEEEdisplaynontitleabstractindextext

%
\IEEEpeerreviewmaketitle

\IEEEraisesectionheading{\section{Introduction}\label{sec:introduction}}
\IEEEPARstart{S}{emantic} segmentation is a computer vision task
in which a trained model classifies pixels
into meaningful high-level categories.
Due to being complementary to object localization,
it represents an important step 
towards advanced image understanding.
Some of the most attractive applications 
include autonomous control \cite{hadsell09jfr}, 
intelligent transportation systems \cite{segvic14mva}, 
assisted photo editing \cite{aksoy18acmtg}
and medical imaging \cite{ronneberger15miccai}.

Early semantic segmentation approaches 
optimized a trade-off between multiple local 
classification cues (texture, color etc) and 
their global agreement across the image \cite{shotton09ijcv}.
Later work improved these ideas with 
non-linear feature embeddings \cite{csurka11ijcv},
multi-scale analysis \cite{farabet13pami}
and depth \cite{kreso16gcpr}.
Spatial consistency has been improved 
by promoting agreement 
between pixels and semantic labels \cite{krahenbuhl11nips},
as well as by learning asymmetric pairwise
semantic agreement potentials \cite{lin16cvpr}.
However, none of these approaches 
has been able to match the improvements
due to deep convolutional models
\cite{lecun90procieee,farabet13pami}.

Deep convolutional models have caused 
an unprecedented rate 
of computer vision development.
Model depth has been steadily increasing 
from 8 levels \cite{krizhevsky12nips} to 19 \cite{simonyan15iclr_a}, 
22 \cite{szegedy15cvpr}, 152 \cite{he16cvpr}, 
201 \cite{huang17cvpr}, and beyond \cite{he16cvpr}.
Much attention has been directed towards 
residual models (also known as ResNets)
\cite{he16cvpr,he16eccv}
in which each processing step 
is expressed as a sum between 
a compound non-linear unit and its input.
This introduces an auxiliary information path 
which allows a direct gradient propagation 
across the layers,
similarly to the flow 
of the state vector across LSTM cells.
However, in contrast 
to the great depth of residual models, 
Veit et al \cite{veit16nips} have empirically determined that 
most training occurs along relatively short paths.
Hence, they have conjectured that a residual model acts 
as an exponentially large ensemble 
of moderately deep sub-models.
This view is especially convincing 
in the case of residual connections 
with identity mappings \cite{he16eccv}.

Recent approaches \cite{larsson17iclr,huang17cvpr}
replicate and exceed the success of residual models
by introducing skip-connections 
across layers.
This encourages feature sharing 
and discourages overfitting
(especially when semantic classes 
 have differing complexities),
while also favouring the gradient flow towards early layers.
Our work is based on densely connected models 
(also known as DenseNets) \cite{huang17cvpr}
in which the convolutional units operate 
on concatenations of all previous features 
at the current resolution.
Our DenseNet-based models for semantic segmentation 
outperform counterparts based on ResNets \cite{he16eccv} 
and more recent dual path networks \cite{chen17nips}.
Another motivation for using DenseNets
is their potential for saving memory 
due to extensive feature reuse \cite{huang17cvpr}.
However, this potential is not easily materialized
since straightforward backprop implementations 
require multiple caching of concatenated features.
We show that these issues can be effectively addressed
by aggressive gradient checkpointing \cite{efficientdensenet17arxiv}
which leads to five-fold memory reduction
with only 20\% increase in training time.

Regardless of the particular architecture,
deep convolutional models for semantic segmentation
must decrease the spatial resolution of deep layers
in order to meet strict GPU memory limitations.
Subsequently, the deep features have to be 
carefully upsampled to the image resolution
in order to generate correct predictions 
at semantic borders and small objects.
Some approaches deal with this issue 
by decreasing the extent of subsampling
with dilated filtering
\cite{zhao17cvpr,chenliangchieh18pami,
  chenliangchieh17arxiv,RotPorKon18a,yang18cvpr}.
Other approaches gradually upsample 
deep convolutional features 
by exploiting cached max-pool switches
\cite{badrinarayanan17pami,noh15iccv}
or activations from earlier layers
\cite{ronneberger15miccai,jegou16corr, ghiasi16eccv,lin17cvpr,peng17cvpr}.
Our approach is related to the latter group
as we also blend the semantics of the deep features
with the location accuracy of the early layers.
However, previous approaches feature 
complex upsampling datapaths 
which require a lot of computational resources.
We show that powerful models can be achieved 
even with minimalistic upsampling,
and that such models are very well suited 
for fast processing of large images.

In this paper, we present an effective lightweight architecture 
for semantic segmentation of large images,
based on DenseNet features and 
ladder-style \cite{valpola14corr} upsampling.
We propose several improvements 
with respect to our previous work \cite{kreso17cvrsuad},
which lead to better accuracy and faster execution 
while using less memory and fewer parameters.
Our consolidated contribution is three-fold.
First, we present an exhaustive study 
of using densely connected \cite{huang17cvpr}
feature extractors for efficient
semantic segmentation.
Second, we propose a lean ladder-style 
upsampling datapath \cite{valpola14corr}
which requires less memory and achieves
a better \mean{IoU}/FLOP trade-off
than previous approaches.
Third, we further reduce the training memory footprint
by aggressive re-computation
of intermediate activations during 
convolutional backprop \cite{efficientdensenet17arxiv}.
The proposed approach strikes an excellent balance
between prediction accuracy and model complexity.
Experiments on Cityscapes, CamVid, ROB 2018 
and Pascal VOC 2012 demonstrate 
state-of-the-art recognition performance and
execution speed with modest training requirements.

\section{Related Work}


Early convolutional models for semantic segmentation
had only a few pooling layers and were trained 
from scratch \cite{farabet13pami}.
Later work built on image classification models 
pre-trained on ImageNet \cite{simonyan15iclr_a,he16cvpr,huang17cvpr},
which typically perform 5 downsamplings before aggregation.
The resulting loss of spatial resolution requires
special techniques for upsampling the features 
back to the resolution of the source image.
Early upsampling approaches were based on
trained filters \cite{shelhamer17pami} 
and cached switches from strided max-pooling layers 
\cite{noh15iccv,badrinarayanan17pami}.
More recent approaches force 
some strided layers to produce non-strided output
while doubling the dilation factor
of all subsequent convolutions
\cite{sermanet14iclr,yu16iclr,chenliangchieh18pami}. 
This decreases the extent of subsampling
while ensuring that the receptive field 
of the involved features
remains the same as in pre-training.

In principle, dilated filtering
can completely recover the resolution  
without compromising
pre-trained parameters.
However, there are two important shortcomings
due to which this technique should be used 
sparingly \cite{chenliangchieh18eccv}
or completely avoided \cite{lintsungyi17cvpr,kreso17cvrsuad}.
First, dilated filtering substantially increases 
computational and GPU memory requirements,
and thus precludes real-time inference
and hinders training on single GPU systems.
Practical implementations alleviate this
by recovering only up to the last two subsamplings,
which allows subsequent inference 
at 8$\times$ subsampled resolution 
\cite{chenliangchieh18pami,zhao17cvpr}.
Second, dilated filtering treats semantic segmentation
exactly as if it were ImageNet classification,
although the two tasks 
differ with respect to location sensitivity.
Predictions of a semantic segmentation model 
must change abruptly in pixels at semantic borders.
On the other hand, image classification predictions
need to be largely insensitive to the location 
of the object which defines the class.
This suggests that optimal semantic segmentation performance 
might not be attainable with architectures
designed for ImageNet classification.

We therefore prefer to keep the downsampling 
and restore the resolution 
by blending semantics of deep features 
with location accuracy 
of the earlier layers \cite{long15cvpr}.
This encourages the deep layers 
to discard location information
and focus on abstract image properties
\cite{valpola14corr}.
Practical realizations avoid 
high-dimensional features
at output resolution \cite{long15cvpr}
by ladder-style upsampling \cite{valpola14corr,ronneberger15miccai,lintsungyi17cvpr}.
In symmetric encoder-decoder approaches,
\cite{badrinarayanan17pami,ronneberger15miccai,jegou16corr}
the upsampling datapath 
mirrors the structure 
of the downsampling datapath.
These methods achieve 
rather low execution speed
due to excessive capacity 
in the upsampling datapath.
Ghiasi et al \cite{ghiasi16eccv} 
blend predictions 
(instead of blending features)
by prefering the deeper layer 
in the middle of the object,
while favouring the earlier layer 
near the object boundary.
Pohlen et al \cite{pohlen17cvpr} propose 
a two-stream residual architecture
where one stream is always at the full resolution, 
while the other stream is first subsampled 
and subsequently upsampled by blending with the first stream.
Lin et al \cite{lin17cvpr} perform the blending
by a sub-model called RefineNet
comprised of 8 convolutional and several other layers
in each upsampling step.
Islam et al \cite{islam17cvpr} blend upsampled predictions
with two layers from the downsampling datapath.
This results in 4 convolutions and one 
elementwise multiplication in each upsampling step.
Peng et al \cite{peng17cvpr} blend predictions
produced by convolutions with very large kernels.
The blending is performed by one 3$\times$3 deconvolution,
two 3$\times$3 convolutions,
and one addition in each upsampling step.
In this work, we argue for
a minimalistic upsampling path
consisting of only one 3$\times$3 convolution
in each upsampling step \cite{kreso17cvrsuad}.
In comparison with symmetric upsampling 
\cite{ronneberger15miccai,badrinarayanan17pami},
this substantially reduces 
the number of 3$\times$3 convolutions 
in the upsampling datapath.
For instance, a VGG 16 feature extractor 
requires 13 3$\times$3 convolutions 
in the symmetric case \cite{badrinarayanan17pami},
but only 4 3$\times$3 convolutions 
in the proposed setup.
To the best of our knowledge, such solution
has not been previously used for semantic segmentation,
although there were uses in object detection \cite{lintsungyi17cvpr}, 
and instance-level segmentation \cite{he17iccv}.
%
%
Additionally, our lateral connections 
differ from 
\cite{ghiasi16eccv,islam17cvpr,peng17cvpr},
since they blend predictions, 
while we blend features.
Blending features improves the modelling power,
but is also more computationally demanding.
We can afford to blend features 
due to minimalistic upsampling
and gradient checkpointing
which will be explained later.
Similarly to \cite{ghiasi16eccv,islam17cvpr,zhao17cvpr}, 
we generate semantic maps at all resolutions
and optimize cross entropy loss 
with respect to all of them.

Coarse spatial resolution 
is not the only shortcoming 
of features designed for 
ImageNet classification.
These features may also have 
insufficient receptive field
to support the correct semantic prediction.
This issue shows up in pixels 
situated at smooth image regions
which are quite common in urban scenes.
Many of these pixels are projected 
from objects close to the camera,
which makes them extremely important 
for high-level tasks
(circumstances 
of the first fatal incident
of a level-2 autopilot are
a sad reminder of this observation).
Some approaches address this issue
by applying 3$\times$3 convolutions 
with very large dilation factors
\cite{chenliangchieh18pami,yu16iclr,chenliangchieh18eccv}.
However, sparse sampling may trigger 
undesired effects due to aliasing.
The receptive field
can also be enlarged by 
extending the depth of the model \cite{islam17cvpr}.
However, the added capacity 
may result in overfitting.
Correlation between 
distant parts of the scene
can be directly modelled 
by introducing long-range connections
\cite{krahenbuhl11nips,lin16cvpr,zhaohengshuang18eccv}.
However, these are often unsuitable due to 
large capacity and computational complexity.
A better ratio between receptive range and complexity
is achieved with spatial pyramid pooling (SPP) \cite{lazebnik06cvpr,he15pami}
which augments the features with their spatial pools 
over rectangular regions of varying size \cite{zhao17cvpr}.
Our design proposes slight improvements 
over \cite{zhao17cvpr}
as detailed in \ref{ssec:spp}.
Furthermore, we alleviate the inability 
of SPP to model spatial layout
by inserting a strided pooling layer
in the middle of the third convolutional block.
This increases the receptive range
and retains spatial layout
without increasing the model capacity.

To the best of our knowledge, 
there are only two published works
on DenseNet-based semantic segmentation
\cite{jegou16corr,yang18cvpr}.
However, these approaches fail to position DenseNet
as the backbone with the largest potential 
for memory-efficient feature extraction\footnote{Here 
  we do not consider 
  reversible models
  \cite{gomez17nips} 
  due to poor availability of
  ImageNet-pretrained parameters, 
  and large computational complexity 
  due to increased dimensionality 
  of the deep layers \cite{jacobsen18iclr}.}.
This potential is caused by a specific design 
which encourages inter-layer sharing \cite{huang17cvpr}
instead of forwarding features across the layers.
Unfortunately, automatic differentiation 
is unable to exploit this potential  
due to concatenation, batchnorm and projection layers.
Consequently, straightforward DenseNet implementations 
actually require a little bit more memory than 
their residual counterparts \cite{kreso17cvrsuad}.
Fortunately, this issue  
can be alleviated with checkpointing 
\cite{chentianqi16arxiv,efficientdensenet17arxiv}.
Previous work on checkpointing segmentation models 
considered only residual models \cite{RotPorKon18a},
and therefore achieved 
only two-fold memory reduction.
We show that DenseNet has much more to gain 
from this technique by achieving 
up to six-fold memory reduction
with respect to the baseline.



\section{Comparison between ResNets and DenseNets}
\label{sec:resdense}

Most convolutional architectures 
\cite{simonyan15iclr_a,szegedy15cvpr,he16eccv,huang17cvpr}
are organized as a succession 
of processing blocks 
which process image representation
on the same subsampling level.
Processing blocks are organized 
in terms of convolutional units
\cite{he16eccv}
which group several convolutions 
operating as a whole.

\begin{figure}[h]
  \begin{center}
    \includegraphics[width=.95\columnwidth]{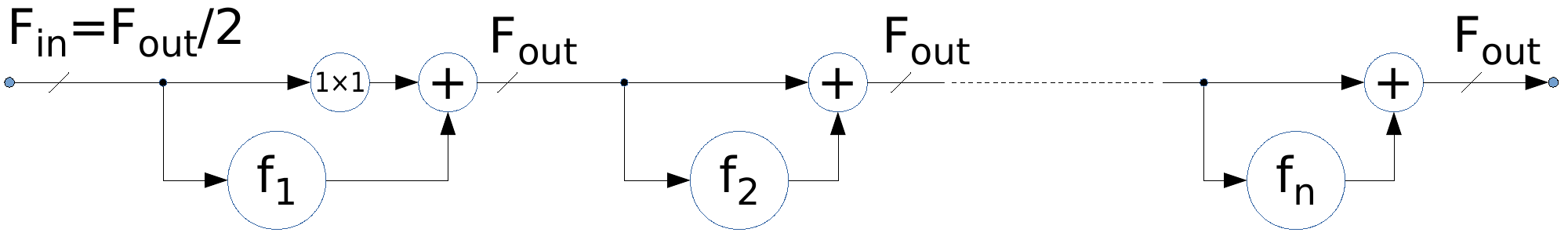}
    
    (a)

    \includegraphics[width=.95\columnwidth]{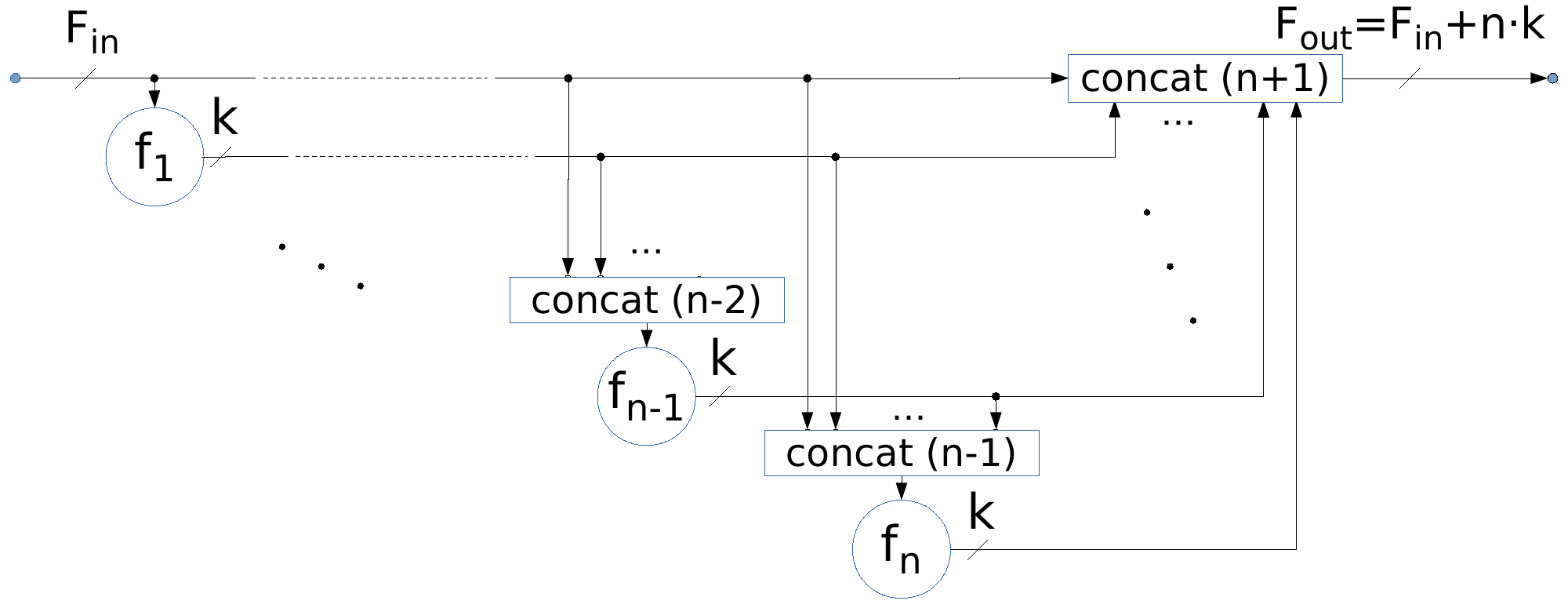}
    
    (b)
  \end{center}
  \caption{%
    A pre-activation residual block 
    \cite{he16eccv} with n units (a) 
    and the corresponding 
    densely connected block 
    \cite{huang17cvpr} (b).
    Labeled circles correspond to 
    convolutional units ($f_1$-$f_k$),
    summations (+) and 
    n-way concatenations - concat (n).
    All connections are 3D tensors 
    D$\times$H$\times$W, 
    where D is designated 
    above the connection line
    (for simplicity, we assume 
    the batch size is 1).
    $F_\mathrm{in}$ and $F_\mathrm{out}$
    denote numbers of feature maps 
    on the processing block 
    input and output, respectively.
  }
  \label{fig:cmpnet}
\end{figure}

We illustrate similarities and differences 
between ResNet and DenseNet architectures 
by comparing the respective processing blocks,
as shown in Figure \ref{fig:cmpnet}.
We consider the most widely used variants:
DenseNet-BC (bottleneck - compression) \cite{huang17cvpr}
and pre-activation ResNets 
with bottleneck \cite{he16eccv}.

\subsection{Organization of the processing blocks}
\label{ss:structure}

The main trait of a residual model
is that the output of each convolutional unit 
is summed with its input 
(cf.\ Figure \ref{fig:cmpnet}(a)).
Hence, all units of a residual block 
have a constant output dimensionality $F_\mathrm{out}$.
Typically, the first ResNet unit
increases the number of feature maps
and decreases the resolution 
by strided projection.
On the other hand, DenseNet units operate 
on a concatenation of the main input with 
all preceding units in the current block
(cf.\ Figure \ref{fig:cmpnet}(b)).
Thus, the dimensionality of a DenseNet block 
increases after each convolutional unit. 
The number of feature maps 
produced by each DenseNet unit 
is called the growth rate
and is defined by the hyper-parameter $k$. 
Most popular DenseNet variations have $k$=32,
however we also consider DenseNet-161 with $k$=48.

In order to reduce the computational complexity, 
both ResNet and DenseNet units 
reduce the number of feature maps
before 3$\times$3 convolutions. 
ResNets reduce the dimensionality to $F_\mathrm{out}/4$, 
while DenseNets go to $4k$.
DenseNet units have two convolutions 
(1$\times$1, 3$\times$3), 
while ResNet units require three convolutions 
(1$\times$1, 3$\times$3, 1$\times$1) 
in order to restore the dimensionality 
of the residual datapath.
The shapes of ResNet convolutions are: 
1$\times$1$\times F_\mathrm{out}\times F_\mathrm{out}$/4,
3$\times$3$\times F_\mathrm{out}/4\times F_\mathrm{out}$/4, and
1$\times$1$\times F_\mathrm{out}/4\times F_\mathrm{out}$.
The convolutions in i-th DenseNet unit are 
1$\times$1$\times[F_\mathrm{in}$+(i-1)$\cdot\mathrm{k}]\times$4k, and
3$\times$3$\times$4k$\times$k.
All DenseNet and pre-activation ResNet units \cite{he16eccv}
apply batchnorm and ReLU activation 
before convolution (BN-ReLU-conv).
On the other hand, the original ResNet design \cite{he16cvpr}
applies convolution at the onset (conv-BN-ReLU),
which precludes negative values 
along the skip connections.

\subsection{Time complexity}
\label{ssec:time}

Both architectures encourage 
exposure of early layers to the loss signal.
However, the distribution of the 
representation dimensionality differs:
ResNet keeps it constant throughout the block, 
while DenseNet increases it towards the end. 
DenseNet units have the following advantages: 
i) producing fewer feature maps 
   (k vs $F_\mathrm{out}$),
ii)
   lower average input dimensionality,
iii)
  fewer convolutions per unit: 2 vs 3.
Asymptotic per-pixel time complexities
of DenseNet and ResNet blocks are respectively:
$O(k^2n^2)$=$O(F_\mathrm{out}^2)$ and
$O(nF_\mathrm{out}^2)$.
Popular ResNets alleviate asymptotic disadvantage
by having only $n$=3 units
in the first processing block
which is computationally the most expensive.
Hence, the temporal complexity 
of DenseNet models is only moderately lower 
than comparable ResNet models,
although the gap increases 
with capacity \cite{huang17cvpr}.

We illustrate this by comparing 
ResNet-50 (26$\cdot$10$^6$ parameters) 
and DenseNet-121 (8$\cdot$10$^6$ parameters).
Theoretically, ResNet-50 requires  
around 33\% more floating operations 
than DenseNet-121 \cite{huang17cvpr}.
In practice the two models achieve 
roughly the same speed on GPU hardware
due to popular software frameworks
being unable to implement concatenations 
without costly copying across GPU memory.
Note however that we have not considered 
improvements based on learned grouped convolutions
\cite{huang18cvpr} and early-exit
classifiers \cite{huang18iclr}.

\subsection{Extent of caching required by backprop}
\label{ssec:caching}

Although DenseNet blocks do not achieve
decisive speed improvement for small models,
they do promise substantial gains 
with respect to training flexibility.
Recall that backprop requires 
caching of input tensors 
in order to be able to calculate gradients
with respect to convolution weights.
This caching may easily 
overwhelm the GPU memory,
especially in the case of 
dense prediction in large images.
The ResNet design implies 
a large number of feature maps 
at input of each convolutional unit
(cf.\,Figure \ref{fig:cmpnet}).
The first convolutional unit incurs 
a double cost despite receiving 
half feature maps due to 
$2^2$ times larger spatial dimensions.
This results in $c_\mathrm{RN}$ 
per-pixel activations which is clearly
$O(n\cdot F_\mathrm{out})$:
\begin{equation}
  c_\mathrm{RN} = (n+1) \cdot F_\mathrm{out} \; .
  \label{eq:crn}
\end{equation}

The DenseNet design alleviates this 
since many convolutional inputs are shared.
An optimized implementation would need to cache
only $c_\mathrm{DN}$ per-pixel activations:
\begin{equation}
c_\mathrm{DN} = 
  F_\mathrm{in} + (n-1) \cdot k \approx F_\mathrm{out} \; .
  \label{eq:cdn}
\end{equation}
Equations (\ref{eq:crn}) and (\ref{eq:cdn}) 
suggest that a DenseNet block has 
a potential for $n$-fold reduction 
of the backprop memory footprint
with respect to a ResNet block
with the same $F_\mathrm{out}$.
We note that ResNet-50 and DenseNet-121 have 
equal $F_\mathrm{out}$
in the first three processing blocks,
while ResNet-50 is twice as large
in the fourth processing block.

Unfortunately, exploiting the described potential
is not straightforward,
since existing autograd implementations 
perform duplicate caching of activations
which pass through multiple concatenations.
Hence, a straightforward implementation
has to cache the output of each unit
in all subsequent units at least two times:
i) as a part of the input to the first batchnorm
ii) as a part of the input to 
the first convolution (1$\times$1).
Due to this, memory requirements of 
unoptimized DenseNet implementations
grow as $O(kn^2)$ \cite{efficientdensenet17arxiv}
instead of $O(kn) = O(F_\mathrm{out})$
as suggested by (\ref{eq:cdn}).
Fortunately, this situation 
can be substantially improved 
with gradient checkpointing.
The idea is to instruct autograd to cache
only the outputs of convolutional units 
as suggested by (\ref{eq:cdn}),
and to recompute the rest during the backprop.
This can also be viewed as a custom backprop step 
for DenseNet convolutional units. 
We postpone the details for Section \ref{sec:ckpt}.


\subsection{Number of parameters}
\label{ssec:params}

It is easy to see that 
the number of parameters 
within a processing block
corresponds to the number of per-pixel multiplications considered in \ref{ssec:time}:
each parameter is multiplied 
exactly once for each pixel.
Hence, the number of DenseNet parameters
is $O(F_\mathrm{out}^2)$,
while the number of ResNet parameters
is $O(n F_\mathrm{out}^2)$.
However, the correspondence between
time complexity and the number of parameters
does not translate to the model level
since time complexity is linear 
in the number of pixels.
Hence, per-pixel multiplications make
a greater contribution to time-complexity  
in early blocks than in later blocks.
On the other hand, 
the corresponding contribution 
to the number of parameters
is constant across blocks.
Therefore, the largest influence 
to the model capacity 
comes from the later blocks 
due to more convolutional units
and larger output dimensionalities.

Table\ \ref{tab:params} compares
the counts of convolutional weights
in ResNet-50 and DenseNet-121.
We see that DenseNet-121 has 
twice as much parameters in block 1,
while the relation is opposite in block 3.
The last residual block has more capacity 
than the whole DenseNet-121.
In the end, DenseNet-121 has three times 
less parameters than ResNet-50. 
\begin{table}[h]
  \caption{Count of convolutional weights 
   across blocks (in millions). 
   ResNet-50 has $n$=$[3,4,6,3]$ and
   $F_\mathrm{out}$=$[256,512,1024,2048]$
   while DenseNet-121 has $n$=$[6,12,24,16]$  and $F_\mathrm{out}$=$[256,512,1024,1024]$. 
  }
  \label{tab:params}
  \footnotesize
  \centering
  \begin{tabular}{l|cccc}
   block@subsampling  
                 & B1@/4 & B2@/8 & B3@/16 & B4@/32 \\
   \hline
   Resnet-50     & 0.2 & 1.2 & 7.1 & 14.9\\
   DenseNet-121  & 0.4 & 1.0 & 3.3 & 2.1 \\
   \hline
  \end{tabular}
\end{table}

\subsection{DenseNets as regularized ResNets}

We will show that each DenseNet block D 
can be realized with 
a suitably engineered ResNet block R
such that each R$_i$ produces
$F_\mathrm{out} = F_\mathrm{in}+nk$ maps.
Assume that R$_i$ is engineered
so that the maps at indices
$F_\mathrm{in} + ik$ through 
$F_\mathrm{in} + (i+1)k$
are determined by non-linear mapping $r_i$,
while all remaining maps are set to zero.
Then, the effect of residual connections
becomes very similar to the concatenation
within the DenseNet block.
Each non-linear mapping $r_i$ 
observes the same input 
as the corresponding DenseNet unit D$_i$:
there are $F_\mathrm{in}$ maps 
corresponding to the block input,
and $(i-1)k$ maps produced by previous units.
Hence, $r_i$ can be implemented by 
re-using weights from D$_i$,
which makes R equivalent to D.

We see that the space of DenseNets 
can be viewed as a sub-space of ResNet models
in which the feature reuse is heavily enforced.
A ResNet block capable to learn
a given DenseNet block requires
an $n$-fold increase in time complexity
since each R$_i$ has to produce
$O(kn)$ instead of $O(k)$ feature maps.
The ResNet block would also require 
an $n$-fold increase of backprop caching
and an $n$-fold increase in capacity
following the arguments in 
\ref{ssec:caching} and \ref{ssec:params}.
We conclude that DenseNets 
can be viewed as strongly regularized ResNets.
Hence, DenseNet models may generalize better
when the training data is scarce.

\begin{figure*}[b]
  \begin{center}
  \includegraphics[width=0.8\textwidth]{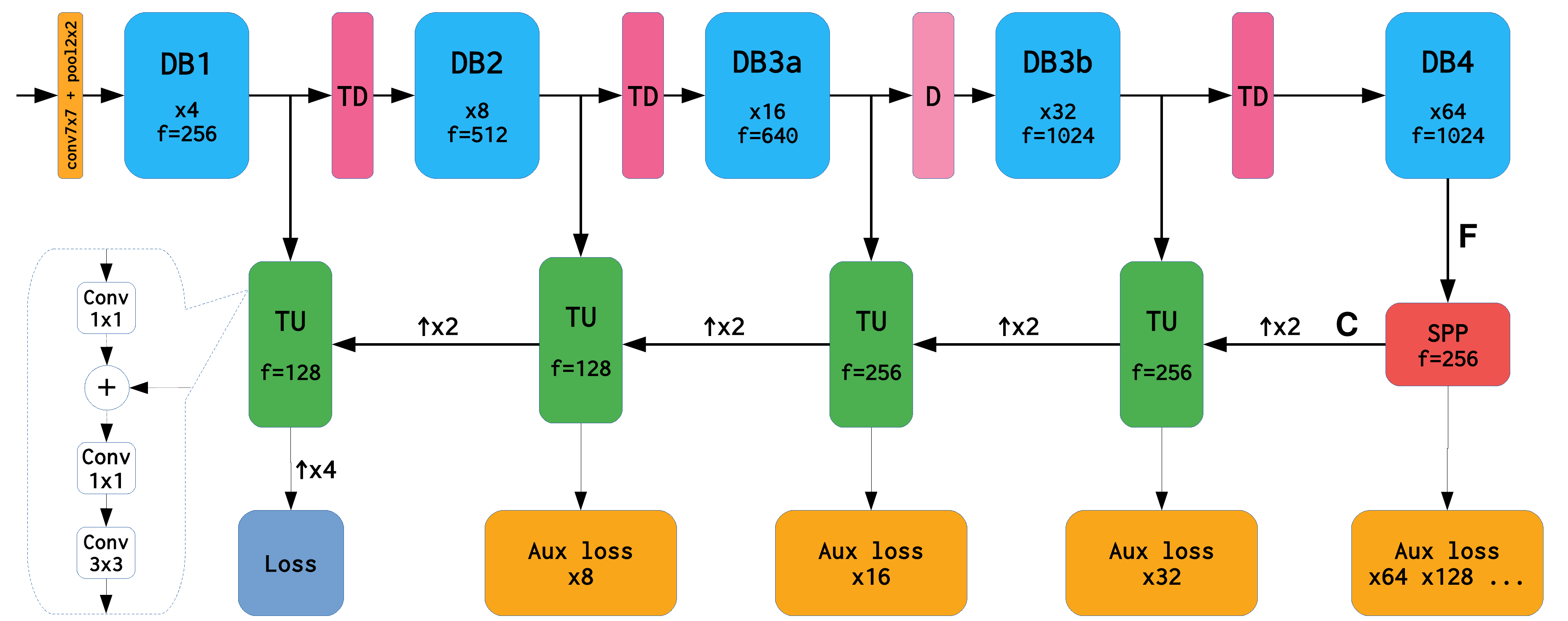}
  \end{center}
   \caption{Architecture of the proposed segmentation model with the
   DenseNet-121 downsampling datapath.
   Each dense block (DBx) is annotated with
   the subsampling factor of the input tensor.
   The number of output feature maps 
   is denoted with f. 
   The transition-up (TU) blocks blend 
   low-resolution high-level features with
   high-resolution low-level features.}
  \label{fig:architecture}
\end{figure*}

\section{The proposed architecture}
\label{sec:arch}

We propose a light-weight semantic segmentation architecture
featuring high accuracy, low memory footprint 
and high execution speed.
The architecture consists of two datapaths
which are designated by two horizontal rails 
in Figure~\ref{fig:architecture}.
The downsampling datapath is composed
of a modified DenseNet feature extractor \cite{huang17cvpr},
and a lightweight 
spatial pyramid pooling module (SPP)
\cite{zhao17cvpr}. 
The feature extractor 
transforms the input image 
into the feature tensor $\mathbf{F}$
by gradually reducing the spatial resolution
and increasing the number of feature maps
(top rail in Figure~\ref{fig:architecture}).
The SPP module enriches the DenseNet features
with context information and creates 
the context-aware features $\mathbf{C}$.
The upsampling datapath transforms 
the low-resolution features $\mathbf{C}$
to high-resolution semantic predictions 
(bottom rail in Figure~\ref{fig:architecture}).
This transformation is achieved by efficient blending 
of semantics from deeper layers 
with fine details from early layers.

\subsection{Feature extraction}

The DenseNet feature extractor 
\cite{huang17cvpr}
consists of dense blocks (DB) 
and transition layers (TD)
(cf.\,Figure~\ref{fig:architecture}).
Each dense block is a concatenation 
of convolutional units, 
while each convolutional unit 
operates on a concatenation 
of all preceding units 
and the block input,
as detailed in Section \ref{sec:resdense}.
Different from the original DenseNet design,
we split the dense block DB3 into two fragments (DB3a and DB3b)
and place a strided average-pooling layer (D) in-between them.
This enlarges the receptive field 
of all convolutions after DB3a,
while decreasing their computational complexity.
In comparison with dilated filtering 
\cite{chenliangchieh18pami}, 
this approach trades-off spatial resolution
(which we later restore with ladder-style blending)
for improved execution speed and reduced memory footprint.
We initialize the DB3b filters with 
ImageNet-pretrained weights
of the original DenseNet model, 
although the novel pooling layer 
alters the features in a way
that has not been seen 
during ImageNet pretraining.
Despite this discrepancy, 
fine-tuning succeeds to recover 
and achieve competitive generalization. 
The feature extractor concludes by concatenating all DB4 units into 
the 64$\times$ subsampled 
representation $\mathbf{F}$.

\subsection{Spatial pyramid pooling}
\label{ssec:spp}

The spatial pyramid pooling module (SPP)
captures wide context information
\cite{lazebnik06cvpr,he15pami,zhao17cvpr}
by augmenting $\mathbf{F}$ 
with average pools over 
several spatial grids.
Our SPP module first projects $\mathbf{F}$ 
to D/2 maps, where D denotes 
the dimensionality of DenseNet features.
The resulting tensor is then average-pooled 
over four grids with 1, 2, 4, and 8 rows.
The number of grid columns is set 
in accordance with the image size 
so that all cells have a square shape.
We project each pooled tensor to D/8 maps
and then upsample with bilinear upsampling. 
We concatenate all results 
with the projected $\mathbf{F}$, 
and finally blend with a 
1$\times$1$\times$D/4 convolution.
The shape of the resulting 
context-aware feature tensor $\mathbf{C}$
is H/64$\times$W/64$\times$D/4.
The dimensionality of $\mathbf{C}$
is 48 times less than 
the dimensionality of the input image
(we assume DenseNet-121, D=1024).

There are two differences 
between our SPP module
and the one proposed in \cite{zhao17cvpr}.
First, we adapt the grid
to the aspect ratio of input features:
each grid cell always averages a square area, 
regardless of the shape of the input image.
Second, we reduce 
the dimensionality of input features 
before the pooling 
in order to avoid 
increasing the output dimensionality.

\subsection{Upsampling datapath}

The role of the upsampling path is to recover 
fine details lost due to the downsampling.
The proposed design is based on
minimalistic transition-up (TU) blocks.
The goal of TU blocks is 
to blend two representations 
whose spatial resolutions
differ by a factor of 2.
The smaller representation comes 
from the upsampling datapath 
while the larger representation comes 
from the downsampling datapath via skip connection.
We first upsample the smaller representation 
with bilinear interpolation 
so that the two representations 
have the same resolution. 
Subsequently, we project 
the larger representation 
to a lower-dimensional space 
so that the two representations 
have the same number of feature maps. 
This balances the relative influence 
of the two datapaths 
and allows to blend the two representations 
by simple summation.
Subsequently, we apply 
a 1$\times$1 convolution 
to reduce the dimensionality (if needed),
and conclude with 3$\times$3 convolution
to prepare the feature tensor 
for subsequent blending.
The blending procedure is 
recursively repeated 
along the upsampling datapath 
with skip-connections 
arriving from outputs of each dense block. The final transition-up block produces logits
at the resolution of the DenseNet stem. 
The dense predictions on input resolution
are finally obtained 
by 4$\times$ bilinear upsampling.

The presented minimalistic design 
ensures fast execution 
due to only one 3$\times$3 convolution 
per upsampling step,
and low memory footprint due to
few convolutions, and
low dimensionality of feature tensors.
The memory footprint can be further reduced 
as detailed in Section \ref{sec:ckpt}.
The proposed upsampling datapath has 
much fewer parameters 
than the downsampling datapath
and therefore discourages overfitting 
to low-level texture 
as we illustrate in the experiments.


\section{Gradient checkpointing}
\label{sec:ckpt}

Semantic segmentation requires 
extraordinary amounts 
of memory during training,
especially on large input resolutions.
These requirements may lead to difficulties
due to strict limitations of GPU RAM.
For example, it is well known that
small training batches may lead
to unstable batchnorm statistics 
and poor learning performance.
This problem can not be overcome
by accumulating backward passes between updates,
and therefore represents a serious obstacle
towards achieving competitive performance.

The extent of backprop-related caching
can be reduced with gradient checkpointing
\cite{chentianqi16arxiv}.
The main idea is to instruct forward pass
to cache only a carefully selected 
subset of all activations.
These activations are subsequently used
for re-computing 
non-cached activations during backprop.
We refer to explicitly cached nodes
of the computational graph 
as gradient checkpoints.
The subgraph between two gradient checkpoints 
is called a checkpointing segment.
The backward pass iterates over 
all checkpointing segments
and processes them as follows.
First, forward pass activations are recomputed
starting from the stored checkpoint. 
Second, the gradients are computed 
via the standard backward pass.
The local cache is released as soon 
as the corresponding segment is processed
i.e.\ before continuing to the next segment.

We note that segment granularity 
affects space and time efficiency. 
Enlarging the checkpoint segments
always reduces the memory footprint 
of the forward pass.
However, the influence to 
backward pass memory requirements
is non-trivial.
Larger segments require more memory 
individually as they need 
to re-compute all required activations
and store them in the local cache.
At some point, we start to lose the gains 
obtained during forward pass.
Our best heuristic was 
to checkpoint only outputs 
from 3$\times$3 convolutions 
as they are the most 
compute-heavy operations.
In other words, we propose to re-compute 
the stem, all projections, all batchnorms 
and all concatenations during the backward pass.
Experiments show that this approach
strikes a very good balance 
between maximum memory allocation 
in forward and backward passes.
The proposed checkpointing strategy
is related to the previous approach \cite{efficientdensenet17arxiv}
which puts a lot of effort 
into explicit management of shared storage.
However, here we show that similar results
can be obtained by relying 
on the standard PyTorch memory manager.
We also show that 
custom backprop operations
can be completely avoided 
by leveraging the standard PyTorch module 
\texttt{torch.utils.checkpoint}.
Finally, we propose to achieve further memory gains 
by caching only outputs of 3$\times$3 convolutions
and the input image.
We achieve that by checkpointing 
the stem, transition-down and transition-up blocks,
as well as DenseNet units as a whole.
To the best of our knowledge,
this is the first account
of applying aggressive checkpointing 
for semantic segmentation.
\section{Experiments}

Most of our experiments target road-driving images, 
since the corresponding applications 
require a very large image resolution 
(subsections \ref{ss:city} and \ref{ss:camvid}).
Cross-dataset generalization experiments
are presented in \ref{ss:domain}.
Subsection \ref{ss:pascal} addresses 
Pascal VOC 2012 \cite{everingham15ijcv} 
as the most popular image segmentation dataset.
We show ablation experiments in \ref{ss:ablation} 
and finally explore 
advantages of checkpointing in \ref{ss:ckpt}. 



\subsection{Training details and notation}
We train our models using AMSGrad 
\cite{kingma14corr,sashank18iclr}
with the initial learning rate $4\cdot{10}^{-4}$.
The learning rate is decreased after each epoch
according to cosine learning rate policy.
We divide the learning rate by 4
for all pre-trained weights.
Batch size is an important hyper-parameter 
of the optimization procedure. 
If we train with batch size 1, 
then the batchnorm statistics
fit exactly the image we are training on.
This hinders learning due to large covariate shift
across different images \cite{szegedy15cvpr}.
We combat this by training on random crops
with batch size 16.
Before cropping, we apply a random flip 
and rescale the image with 
a random factor between 0.5 and 2.
The crop size is set to 448, 512 or 768
depending on the resolution of the dataset.
If a crop happens to be larger 
than the rescaled image, then
the undefined pixels are filled 
with the mean pixel.
We train for 300 epochs unless 
otherwise stated.
We employ multiple cross entropy losses 
along the upsampling path
as shown in Figure~\ref{fig:architecture}.
Auxiliary losses use soft targets 
determined as the label distribution
in the corresponding N$\times$N window 
where N denotes the downsampling factor.
We apply loss after each upsampling step, 
and to each of the four pooled tensors 
within the SPP module.
The loss of the final predictions
is weighted by $0.6$,
while the mean auxiliary loss 
contributes with factor $0.4$.
After the training, we recompute 
the batchnorm statistics as exact averages 
over the training set 
instead of decayed moving averages
used during training. 
This practice slightly improves 
the model generalization.

We employ the following notation
to describe our experiments
throughout the section.
LDN stands for Ladder DenseNet, 
the architecture proposed in Section \ref{sec:arch}.
The symbol $d$\upmark$u$ denotes a 
downsampling path which reduces 
the image resolution $d$ times,
and a ladder-style upsampling path which
produces predictions subsampled $u$ times
with respect to the input resolution.
For example, LDN121 64\upmark4 
denotes the model shown in
Figure~\ref{fig:architecture}.
Similarly, DDN and LDDN denote a dilated DenseNet, 
and a dilated DenseNet
with ladder-style upsampling.
The symbol $d\!\downarrow$ denotes a model 
which reduces the image resolution $d$ times
and has no upsampling path.
MS denotes multi-scale evaluation 
on 5 scales (0.5, 0.75, 1, 1.5 and 2),
and respective horizontal flips.
\textoverline{IoU} and Cat.\,\textoverline{IoU}
denote the standard mean IoU metric 
over classes and categories.
The instance-level mean IoU 
(\textoverline{iIoU}) metric
\cite{cordts15cvpr}
emphasizes contribution of pixels 
at small instances,
and is therefore evaluated 
only on 8 object classes. 
The model size is expressed 
as the total number of parameters 
in millions (M).
FLOP denotes the number of fused multiply-add
operations required for inference
on a single 1024$\times$1024 (1MPx) image.

\subsection{Cityscapes}
\label{ss:city}
The Cityscapes dataset contains 
road driving images 
recorded in 50 cities 
during spring, summer and autumn.
The dataset features 19 classes,
good and medium weather, 
large number of dynamic objects, 
varying scene layout and varying background.
We perform experiments on 
5000 finely annotated images
divided into
2975 training,
500 validation,
and 1525 test images. 
The resolution of all images is 1024$\times$2048.

Table \ref{tab:city_val_arch} validates 
several popular backbones 
coupled with the same SPP and upsampling modules.
Due to hardware constraints,
here we train and evaluate on
half resolution images,
and use 448$\times$448 crops.
The first section of the table presents
the DN121 32$\downarrow$ baseline.
The second section presents our models
with ladder-style upsampling.
The LDN121 64\upmark4 model
outperforms the baseline 
for 10 percentage points 
(pp) of \mean{IoU} improvement.
Some improvements occur on small instances 
as a result of finer output resolution
due to blending with low-level features.
Other improvements occur on large instances
since increased subsampling (64 vs 32)
enlarges the spatial context.
Note that LDN121 32\upmark4 slightly outperforms
LDN121 64\upmark4 at this resolution
due to better accuracy at semantic borders.
However, the situation will be opposite
in full resolution images due to larger objects 
(which require a larger receptive field)
and off-by-one pixel annotation errors.
The LDN169 32\upmark4 model 
features a stronger backbone,
but obtains a slight deterioration (0.8pp)
with respect to LDN121 32\upmark4. 
We conclude that half resolution images
do not contain enough training pixels
to support the capacity of DenseNet-169.
The third section demonstrates 
that residual and DPN backbones 
achieve worse generalization 
than their DenseNet counterparts.
The bottom section shows that further upsampling 
(LDN121 32\upmark2)
doubles the computational complexity
while bringing only a slight accuracy improvement.



\begin{table}[h]
\renewcommand{\arraystretch}{1.3}
\caption{Validation of backbone architectures 
  on Cityscapes val.
  Both training and evaluation images 
  were resized to 1024$\times$512.
}
\label{tab:city_val_arch}
\footnotesize
\centering
\begin{tabular}{l|c|c|c|c|c}
   & \multicolumn{2}{c|}{Class} & Cat. & Model & FLOP \\
  Method & \mean{IoU} & \mean{iIoU} & \mean{IoU} & size & 1MPx \\
  \hline
  \hline
  DN121 32$\downarrow$ & 66.2 & 46.7 & 78.3 & 8.2M & 56.1G \\
  \hline
  LDN121 64\upmark4 & 75.3 & 54.8 & 88.1 & 9.5M & 66.5G \\
  LDN121 32\upmark4 & 76.6 & 57.5 & 88.6 & 9.0M & 75.4G \\
  LDN169 32\upmark4 & 75.8 & 55.5 & 88.4 & 15.6M & 88.8G \\
  \hline
  ResNet18 32\upmark4 & 70.9 & 49.7 & 86.7 & 13.3M & 55.7G \\
  ResNet101 32\upmark4 & 73.7 & 54.3 & 87.8 & 45.9M & 186.7G \\
  ResNet50 32\upmark4 & 73.9 & 54.2 & 87.8 & 26.9M & 109.0G \\
  DPN68 32\upmark4 & 74.0 & 53.0 & 87.8 & 13.7M & 59.0G \\

  \hline
  LDN121 32\upmark2 & \textbf{77.5} & \textbf{58.9} & 89.3 & 9.4M & 154.5G \\
  \hline
\end{tabular}
\end{table}

Table \ref{tab:city_val_dilation} 
addresses models which
recover the resolution loss 
with dilated convolutions. 
The DDN-121 8$\downarrow$ model
removes the strided pooling layers
before the DenseNet blocks DB3 and DB4,
and introduces dilation 
in DB3 (\lstinline{rate=2})
and DB4 (\lstinline{rate=4}). 
The SPP output is now 8$\times$ downsampled. 
From there we produce logits and finally 
restore the input resolution 
with bilinear upsampling.
The LDDN-121 8\upmark4 model continues 
with one step of ladder-style upsampling 
to obtain 4$\times$ downsampled predictions 
as in previous LDN experiments.
We observe a 3pp \mean{IoU} improvement
due to ladder-style upsampling. 
The LDDN-121 16\upmark4 model
dilates only the last dense block 
and performs two steps 
of ladder-style upsampling.
We observe a marginal improvement which,
however, still comes short 
of LDN121 32\upmark4
from Table \ref{tab:city_val_arch}.
Training the DDN-121 4$\downarrow$ model 
was infeasible due to huge 
computational requirements 
when the last three blocks operate 
on 4$\times$ subsampled resolution.
A comparison of computational complexity reveals
that the dilated LDDN-121 8\upmark4 model 
has almost 3$\times$ more FLOPs 
than LDN models with similar \mean{IoU} performance.
Finally, our memory consumption measurements
show that LDDN-121 8\upmark4 consumes
around 2$\times$ more GPU memory 
than LDN121 32\upmark4.
We conclude that dilated models
achieve a worse generalization
than their LDN counterparts
while requiring more computational power.

\begin{table}[h]
\renewcommand{\arraystretch}{1.3}
\caption{Validation of dilated models
  on Cityscapes val.
  Both training and evaluation images 
  were resized to 1024$\times$512.}
\label{tab:city_val_dilation}
\footnotesize
\centering
\begin{tabular}{l|c|c|c|c|c}
   & \multicolumn{2}{c|}{Class} & Cat. & Model & FLOP \\
  Method & \mean{IoU} & \mean{iIoU} & \mean{IoU} & size & 1MPx \\
  \hline\hline
  DDN-121 8$\downarrow$ & 72.5 & 52.5 & 85.5 & 8.2M & 147.8B \\
  \hline
  LDDN-121 8\upmark4 & 75.5 & 55.3 & 88.3 & 8.6M & 174.8B \\
  LDDN-121 16\upmark4 & 75.8 & 55.9 & 88.4 & 8.9M & 87.0B \\
  \hline
\end{tabular}
\end{table}

Table \ref{tab:city_val_full} shows 
experiments on full Cityscapes val images
where we train on 768$\times$768 crops. 
We obtain the most interesting results 
with the LDN121 64\upmark4 model
presented in Figure~\ref{fig:architecture}:
79\% \mean{IoU} with a single forward pass 
and 80.3\% with multi-scale (MS) inference.
Models with stronger backbones 
(DenseNet-169, DenseNet-161) 
validate only slightly better.
We explain that by 
insufficient training data
and we expect that successful models need
less capacity for Cityscapes than for
a harder task of discriminating ImageNet classes.

\begin{table}[h]
\renewcommand{\arraystretch}{1.3}
\caption{Validation of various design options
  on full-resolution Cityscapes val.}
\label{tab:city_val_full}
\footnotesize
\centering
\begin{tabular}{l|c|c|c|c}
   & Class & Cat. & Model & FLOP \\
  Method & \mean{IoU} & \mean{IoU} & size & 1MPx \\
  \hline\hline
  LDN121 32\upmark4 & 78.4 & 90.0 & 9.0M & 75.4G \\
  LDN121 64\upmark4 & 79.0 & 90.3 & 9.5M & 66.5G \\
  LDN121 128\upmark4 & 78.4 & 90.1 & 9.9M & 66.2G \\
  \hline
  LDN161 64\upmark4 & 79.1 & 90.2 & 30.0M & 138.7G \\
  \hline
  LDN121 64\upmark4 MS & 80.3 & 90.6 & 9.5M & 536.2G \\
\hline
\end{tabular}
\end{table}

\begin{figure}[h]
  \begin{center}
  \includegraphics[width=\columnwidth]{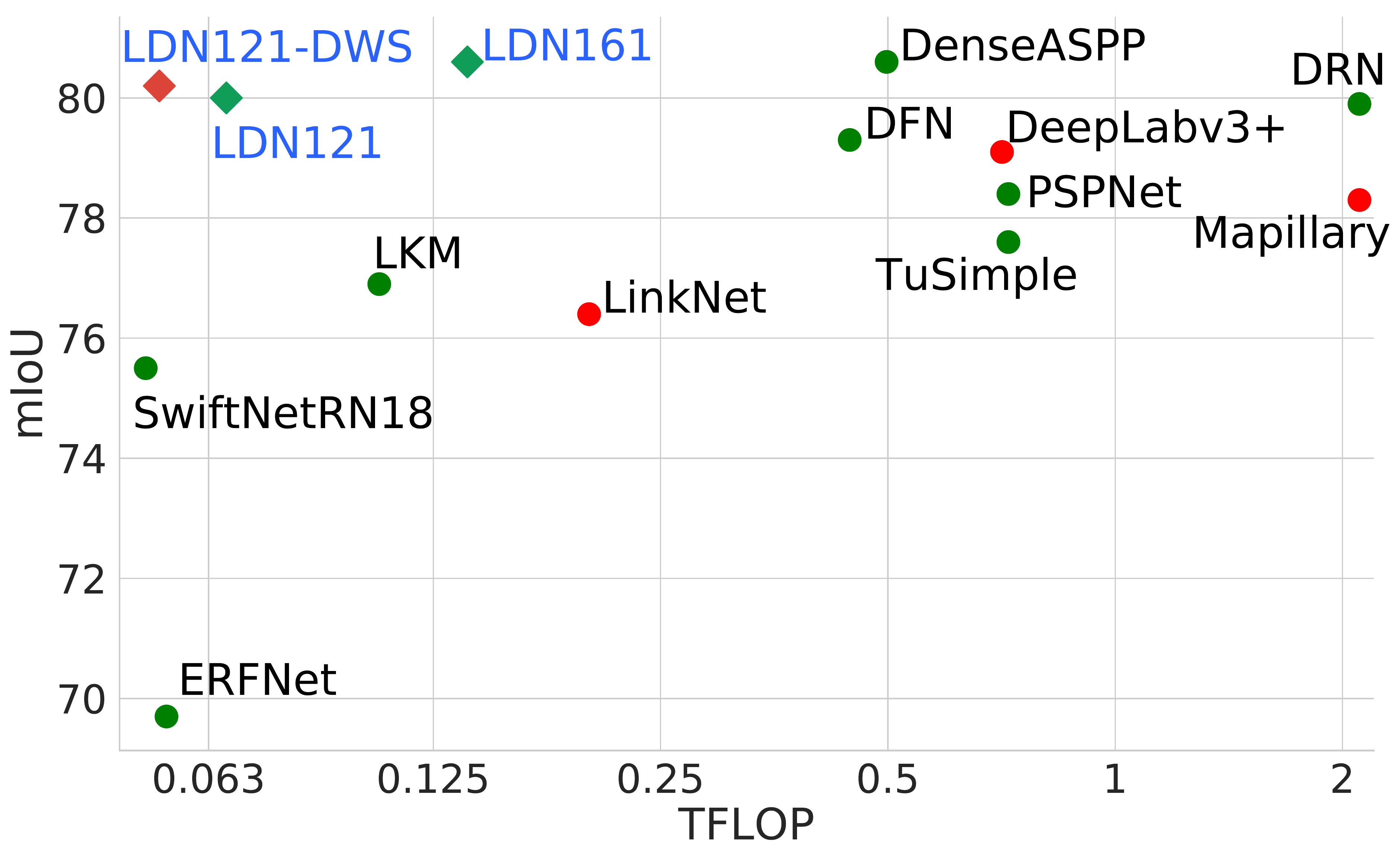}
  \end{center}
  \caption{
    Accuracy vs forward pass complexity 
    on Cityscapes test (green) and val (red)
    for approaches from Table \ref{tab:city_test}.
    All models have been trained 
    only on finely annotated images.
    LDN121 is the first method
    to achieve 80\% \mean{IoU}
    while being applicable 
    in real-time.
  }
  \label{fig:flops_cmp}
\end{figure}

Table \ref{tab:city_test} compares
the results of two of our best models
with the state-of-the-art
on validation and test sets.
All models from the table have been trained
only on finely annotated images.
The label DWS denotes 
depthwise separable convolutions
in the upsampling path 
(cf.\,Table \ref{tab:city_ablation}).
Our models generalize better than 
or equal to all previous approaches, 
while being much more efficient.
In particular, we are the first 
to achieve 80\% \mean{IoU} on Cityscapes test fine 
with only 66.5\,GFLOP per MPx.
Figure \ref{fig:flops_cmp} 
plots the best performing models 
from Table \ref{tab:city_test}
in (\mean{IoU},TFLOP) coordinates.
\begin{table}[h]
\renewcommand{\arraystretch}{1.3}
\caption{Comparison of our two best models 
  with the state-of-the-art 
  on Cityscapes val and test.
  All models have been trained 
  only on finely annotated images.
  For models marked with '\dag' 
  we estimate a lower FLOP-bound 
  by measuring the complexity of the backbone.
  Most of the \mean{IoU} results use multi-scale inference
  while we always show required FLOPs for single-scale inference.
}
\label{tab:city_test}
\footnotesize
\centering
\begin{tabular}{l|c|c|c|c}
   & & Val & Test & FLOP \\
  Method & Backbone & \mean{IoU} & \mean{IoU} & 1MPx\\
  \hline\hline
  ERFNet \cite{romera18tits} & Custom 8$\times$ & 71.5 & 69.7 & 55.4G \\
  SwiftNet \cite{orsic19cvpr} & RN18 32$\times$ & 75.4 & 75.5 & 52.0G \\
  LinkNet \cite{chaurasia17vcip} & RN18 32$\times$ & 76.4 & n/a & 201G \\
  LKM \cite{peng17cvpr} & RN50 32$\times$ & 77.4 & 76.9 & 106G$^{\dag}$ \\
  TuSimple \cite{wang17corr} & RN101 D8$\times$ & 76.4 & 77.6 & 722G$^{\dag}$ \\
  SAC-multiple \cite{zhang17iccv} & RN101 D8$\times$ & 78.7 & 78.1 & 722G$^{\dag}$ \\
  WideResNet38 \cite{wu16corr} & WRN38 D8$\times$ & 77.9 & 78.4 & 2106G$^{\dag}$ \\
  PSPNet \cite{zhao17cvpr} & RN101 D8$\times$ & n/a & 78.4 & 722G$^{\dag}$ \\
  Multi Task \cite{kendall18cvpr} & RN101 D8$\times$ & n/a & 78.5 & 722G$^{\dag}$ \\
  TKCN \cite{wu18arxiv} & RN101 D8$\times$ & n/a & 79.5 & 722G$^{\dag}$ \\
  DFN \cite{yu18cvpr} & RN101 32$\times$ & n/a & 79.3 & 445G$^{\dag}$ \\
  Mapillary \cite{RotPorKon18a} & WRN38 D8$\times$ & 78.3 & n/a & 2106G$^{\dag}$ \\
  DeepLab v3 \cite{chenliangchieh17arxiv} & RN101 D8$\times$ & 79.3 & n/a & 722G$^{\dag}$ \\
  DeepLab v3+ \cite{chenliangchieh18eccv} & X-65 D8$\times$ & 79.1 & n/a & 708G  \\
  DeepLab v3+ \cite{chenliangchieh18eccv} & X-71 D8$\times$ & 79.5 & n/a & n/a  \\
  DRN \cite{zhuang18ICIP} & WRN38 D8$\times$ & 79.7 & 79.9 & 2106G$^{\dag}$ \\
  DenseASPP \cite{yang18cvpr} & DN161 D8$\times$ & 78.9 & $\mathbf{80.6}$ & 498G$^{\dag}$ \\
  \hline
  LDN121 DWS & DN121 64$\times$ & 80.2 & n/a & $\mathbf{54.2}$G \\
  LDN121 64\upmark4 & DN121 64$\times$ & 80.3 & 80.0 & $\mathbf{66.5}$G \\
  LDN161 64\upmark4 & DN161 64$\times$ & $\mathbf{80.7}$ & $\mathbf{80.6}$ & 139G \\
\end{tabular}
\end{table}
The figure clearly shows that our models 
achieve the best trade-off between 
accuracy and computational complexity.


%



\subsection{CamVid}
\label{ss:camvid}

The CamVid dataset contains images of
urban road driving scenes.
We use the 11-class split from 
\cite{badrinarayanan17pami},
which consists of 367 training,
101 validation, and 
233 testing images.
The resolution of all images 
is 720$\times$960.
Following the common practice, 
we incorporate the val subset into train
because it is too small and too easy 
to be useful for validation.
We train all models 
from random initialization (RI),
and by fine-tuning the parameters 
pre-trained on ImageNet (PT).
We train on 512$\times$512 crops
for 400 epochs with pre-training,
and 800 epochs with random init.
All other hyperparameters are the same 
as in Cityscapes experiments.

Table \ref{tab:camvid} shows our results 
on full-resolution CamVid test.
The conclusions are similar 
as on half-resolution Cityscapes val 
(cf. Table \ref{tab:city_val_arch}),
which does not surprise us 
due to similar image resolutions.
LDN121 32\upmark4 wins both in the
pre-trained and in the random init case,
with LDN121 64\upmark4 being the runner-up.
\begin{table}[h]
\renewcommand{\arraystretch}{1.3}
\caption{Single-scale inference on 
  full-resolution CamVid test with 
  ImageNet pre-training (PT)
  and random initialization (RI).
}
\label{tab:camvid}
\footnotesize
\centering
    \begin{tabular}{l|c|c|c|c}
      & PT & RI & Model & FLOP \\
      Method & \mean{IoU} & \mean{IoU} & size & 1MPx \\
      \hline\hline

    LDN121 32\upmark4 & $\mathbf{77.3}$ & $\mathbf{70.9}$ & 9.0M & 75.4G \\
    LDN121 64\upmark4 & 76.9 & 68.7 & 9.5M & 66.5G \\
      \hline
    ResNet18 32\upmark4 & 73.2 & 70.0 & 13.3M & 55.7G \\
    ResNet50 32\upmark4 & 76.1 & 69.9 & 26.9M & 109.0G \\
    ResNet101 32\upmark4& 76.7 & 69.4 & 45.9M & 186.7G \\
    \hline
    \end{tabular}
\end{table}
Table \ref{tab:camvid_others} 
compares our best results 
with the related work on CamVid test
where, to the best of our knowledge, 
we obtain state-of-the-art results.

\begin{table}[h]
\renewcommand{\arraystretch}{1.3}
\caption{Comparison of our models 
  with the state-of-the-art 
  on CamVid test.
  We use multi-scale inference 
  in experiments on full resolution.
}
\label{tab:camvid_others}
\footnotesize
\centering
\begin{tabular}{l|c|c|c|c}
  Method & Backbone & ImgNet & Resolution & \mean{IoU} \\
  \hline\hline      
  Tiramisu \cite{jegou16corr} &
    DenseNet & 
               & half & 66.9 \\
  FC-DRN \cite{casanova18arxiv} &
    DenseResNet &
               & half & 69.4 \\
  \hline
  G-FRNet \cite{gfrnet18arxiv} & 
    VGG-16 &
    \checkmark & half & 68.8 \\
  BiSeNet \cite{bisenet18eccv} &
    Xception39 &
    \checkmark & full & 65.6 \\
  ICNet \cite{icnet18eccv} & 
    ResNet-50 &
    \checkmark & full & 67.1 \\
  BiSeNet \cite{bisenet18eccv} &
    ResNet-18 &
    \checkmark & full & 68.7 \\
  \hline
  \hline
  LDN121 16\upmark2 & DenseNet & 
              & half &  $\mathbf{69.5}$ \\
  LDN121 32\upmark4 & DenseNet & 
               & full & $\mathbf{71.9}$ \\
  LDN121 16\upmark2 & DenseNet &
    \checkmark & half & $\mathbf{75.8}$ \\
  LDN121 32\upmark4 & DenseNet &
    \checkmark & full & $\mathbf{78.1}$ \\
  \hline
\end{tabular}
\end{table}

\subsection{Cross-dataset generalization}
\label{ss:domain}

We explore the capability
of our models to generalize
across related datasets.
We train on Cityscapes and Vistas,
and evaluate on 
Cityscapes, Vistas and KITTI.
Mapillary Vistas \cite{neuhold17iccv} 
is a large road driving dataset
featuring five continents
and diverse lighting, 
seasonal and weather conditions.
It contains 18000 training,
2000 validation, and 5000 test images.
In our experiments, we remap annotations 
to 19 Cityscapes classes, and 
resize all images to width 2048.
The KITTI dataset contains
road driving images
recorded in Karlsruhe
\cite{kitti_semantic}.
It features the Cityscapes 
labeling convention
and depth reconstruction groundtruth.
There are 200 training and
200 test images.
All images are
370$\times$1226.

Table \ref{tab:city_vistas} 
shows that training only on Cityscapes
results in poor generalization
due to urban bias 
and constant acquisition setup. 
On the other hand, Vistas is much more diverse.
Hence, the model trained on Vistas 
performs very well on Cityscapes,
while training on both datasets 
achieves the best results.


\begin{table}[h]
\renewcommand{\arraystretch}{1.3}
\caption{Cross-dataset evaluation
  on half-resolution images. 
  We train separate models on Cityscapes,
  Vistas and their union, 
  and show results on validation sets
  of the three driving datasets.
  Note that this model
  differs from 
  Figure~\ref{fig:architecture}
  since it splits DB4
  instead of DB3.
}
\label{tab:city_vistas}
\footnotesize
\centering
\begin{tabular}{l|c|c|c|c}
  & Training & 
  Cityscapes & Vistas & KITTI \\
  Method & dataset &
    \mean{IoU} & \mean{IoU} & \mean{IoU}
  \\
  \hline\hline
  LDN121 64\upmark4 s4& Cityscapes & 
    76.0 & 44.0 & 59.5 \\
  LDN121 64\upmark4 s4& Vistas & 
    68.7 & 73.0 & 64.7 \\
  LDN121 64\upmark4 s4& Cit.\,+\,Vis. & 
    \textbf{76.2} & \textbf{73.9} & \textbf{68.7} \\
  \hline
\end{tabular}
\end{table}


We briefly describe our submission
\cite{kreso18rob}
to the Robust vision workshop 
held in conjunction with CVPR 2018.
The ROB 2018 semantic segmentation challenge 
features 1 indoor (ScanNet),
and 3 road-driving 
(Cityscapes, KITTI, WildDash) datasets.
A qualifying model has to be trained 
and evaluated on all four benchmarks,
and it has to predict at least 39 classes:
19 from Cityscapes and 20 from ScanNet.
We have configured the LDN169 64\upmark4 model
accordingly, 
and trained it on Cityscapes train+val,
KITTI train, WildDash val and 
ScanNet train+val. 
Our submission has received the runnner-up prize.
Unlike the winning submission
\cite{RotPorKon18a}, 
our model was trained only
on the four datasets provided by the challenge, 
without using any additional 
datasets like Vistas.





\subsection{Pascal VOC 2012}
\label{ss:pascal}

PASCAL VOC 2012 contains
photos from private collections.
There are 6 indoor classes,
7 vehicles, 7 living beings,
and one background class. 
The dataset contains 
1464 train, 1449 validation
and 1456 test images of variable size.
Following related work, we augment the train set
with extra annotations from \cite{hariharan11iccv},
resulting in 10582 training images in augmented set (AUG).
Due to annotation errors in the AUG labels,
we first train for 100 epochs on the AUG set,
and then fine-tune for another 100 epochs 
on train (or train+val).
We use 512$\times$512 crops and divide the learning
rate of pretrained weights by 8. 
All other hyperparameters are 
the same as in Cityscapes experiments.
Table \ref{tab:voc2012} shows that our models set 
the new state-of-the-art on VOC 2012 
without pre-training on COCO.



\begin{table}[h]
\renewcommand{\arraystretch}{1.3}
\caption{Experimental evaluation 
  on Pascal VOC 2012 validation and test.}
\label{tab:voc2012}
\footnotesize
\centering
\begin{tabular}{l|c|c|c|c}
    & & & Val & Test \\
  Method & AUG & MS & \mean{IoU} & \mean{IoU} \\
  \hline
  DeepLabv3+ Res101 & \checkmark & \checkmark & 80.6 & n/a \\
  DeepLabv3+ Xcept & \checkmark & \checkmark & 81.6 & n/a \\
  DDSC \cite{bilinski18cvpr} & \checkmark & & n/a & 81.2 \\
  AAF \cite{ke18eccv} & \checkmark & \checkmark & n/a & 82.2 \\
  PSPNet \cite{zhao17cvpr} & \checkmark & \checkmark & n/a & 82.6 \\
  DFN \cite{yu18cvpr} & \checkmark & \checkmark & 80.6 & 82.7 \\
  EncNet \cite{zhang18cvpr} & \checkmark & \checkmark & n/a & 82.9 \\
  \hline
  LDN121 32\upmark4 & & & 76.4 & n/a  \\
  LDN169 32\upmark4 & \checkmark & \checkmark & 80.5 & 81.6 \\
  LDN161 32\upmark4 & & & 78.6 & n/a \\
  LDN161 32\upmark4 & \checkmark & & 80.4 & n/a  \\
  LDN161 32\upmark4 & \checkmark & \checkmark & $\mathbf{81.9}$ & $\mathbf{83.6}$ \\
  \hline
\end{tabular}
\end{table}


\subsection{Ablation experiments on Cityscapes}
\label{ss:ablation}

Table \ref{tab:city_ablation} evaluates 
the impact of auxiliary loss, SPP,
and depthwise separable convolutions 
on generalization accuracy.
The experiment labeled NoSPP 
replaces the SPP module with 
a single 3$\times$3 convolution.
The resulting 1.5pp performance drop
suggests that SPP brings improvement 
even with 64 times subsampled features.
The subsequent  experiment 
shows that the SPP module 
proposed in \cite{zhao17cvpr}
does not work well with our training on Cityscapes.
We believe that the 1.4pp performance drop
is due to inadequate pooling grids and 
larger feature dimensionality 
which encourages overfitting.
%
The NoAux model applies the loss 
only on final predictions.
The resulting 1.2pp performance hit 
suggests that auxiliary loss 
succeeds to reduce overfitting
on low-level features
within the upsampling path.
The DWS model reduces 
the computational complexity
by replacing all 3$\times$3 convolutions 
in the upsampling path with 
depthwise separable convolutions.
This improves efficiency while
\begin{table}[h]
\renewcommand{\arraystretch}{1.3}
\caption{Impact of auxiliary loss, SPP, 
  and depthwise separable convolutions 
  on generalization accuracy 
  on full-resolution Cityscapes val.}
\label{tab:city_ablation}
\footnotesize
\centering
\begin{tabular}{l|c|c|c}
   & & Model & FLOP \\
  Method & \mean{IoU} & size & 1MPx \\
  \hline\hline
  LDN121 64\upmark4 NoSPP & 77.5 & 10.2M & 66.7G \\
  LDN121 64\upmark4 SPP \cite{zhao17cvpr}& 77.6 & 10.6M & 66.9G \\
  \hline
  LDN121 64\upmark4 NoAux & 77.8 & 9.5M & 66.5G \\
  LDN121 64\upmark4 DWS & 78.6 & 8.7M & 54.2G \\
  \hline
  LDN121 64\upmark4 & $\mathbf{79.0}$ & 9.5M & 66.5G \\
\hline
\end{tabular}
\end{table}
only marginally decreasing accuracy.

Table \ref{tab:city_val_full_jitter} 
shows ablation experiments which evaluate 
the impact of data augmentations 
on generalization.
We observe that random image flip, crop, 
and scale jitter improve the results
by almost 5pp \mean{IoU} and conclude that 
data augmentation is of great importance
for semantic segmentation.

\begin{table}[h]
\renewcommand{\arraystretch}{1.3}
\caption{Impact of data augmentation 
  to the segmentation accuracy (\mean{IoU})
  on Cityscapes val
  while training LDN121 64\upmark4 
  on full images.
}
\label{tab:city_val_full_jitter}
\footnotesize
\centering
\begin{tabular}{l|c|c|c|c}
  augmentation: & none & flip & flip/crop & 
  flip/crop/scale\\
  \hline
   accuracy (\mean{IoU}): & 74.0 & 75.7 & 76.7 & 79.0
\end{tabular}
\end{table}

\subsection{Gradient checkpointing}
\label{ss:ckpt}

Table~\ref{tab:checkpoint} explores effects 
of several checkpointing strategies 
to the memory footprint and the execution speed 
during training of the default LDN model 
(cf.\,Figure \ref{fig:architecture})
on 768$\times$768 images.
\begin{table}[b]
\renewcommand{\arraystretch}{1.3}
\caption{Impact of checkpointing 
  to memory footprint and training speed.
  We train LDN 32\upmark4
  on 768$\times$768 images
  on a Titan Xp with 12 GB RAM.
  Column 2 (Memory) assumes batch size 6. 
}
\label{tab:checkpoint}
\footnotesize
\centering
\begin{tabular}{l|c|c|c}
   & Memory & Max & Train \\
  Checkpointing variant & bs=6 (MB) & BS & FPS \\
  \hline\hline
  baseline - no ckpt & 11265 & 6 & 11.3 \\
  (3x3) & 10107 & 6 & 10.5 \\
  (cat 1x1) & 6620 & 10 & 10.4 \\
  (cat 1x1) (3x3) & 5552 & 12 & 9.7 \\
  (block) (stem) (TD) (UP) & 3902 & 16 & 8.4 \\
  (cat 1x1 3x3) & 3620 & 19 & 10.1 \\
  (cat 1x1 3x3) (stem) (TD) (UP) & 2106 & 27 & 9.2 \\
  \hline
\end{tabular}
\end{table}
We first show the maximum memory allocation 
at any point in the computational graph
while training with batch size 6.
Subsequently, we present the maximum batch size 
we could fit into GPU memory and 
the corresponding training speed 
in frames per second (FPS).
We start from the straightforward baseline
and gradually introduce more and more 
aggressive checkpointing.
The checkpointing approaches 
are designated as follows.
The label \emph{cat} refers to the concatenation 
before a DenseNet unit
(cf.\,Figure \ref{fig:cmpnet}).
The labels 1$\times$1 and 3$\times$3
refer to the first and the second 
BN-ReLu-conv group within a DenseNet unit.
The label \emph{stem} denotes the
7$\times$7 convolution 
at the very beginning of DenseNet \cite{huang17cvpr},
including the following batchnorm, 
ReLU and max-pool operations.
Labels TD and TU correspond to 
the transition-down and 
the transition-up blocks.
The label \emph{block} refers to the entire processing block
(this approach is applicable to most backbones)
Parentheses indicate the checkpoint segment. 
For example, (cat 1$\times$1 3$\times$3) 
means that only inputs to the concatenations
in front of the convolutional unit are cached,
while the first batchnorm, 
the 1$\times$1 convolution
and the second batchnorm 
are re-computed during backprop.
On the other hand, 
(cat 1$\times$1) (3$\times$3) 
means that each convolution 
is in a separate segment.
Here we cache the input to the concatenations
and the input to the second batchnorm,
while the two batchnorms are recomputed.
Consequently, training with 
(cat 1$\times$1 3$\times$3) 
requires less caching and is therefore 
able to accommodate larger batches.

Now we present the most important
results from the table.
The fourth row of the table shows that
checkpointing the (cat 1$\times$1) subgraph
brings the greatest savings 
with respect to the baseline (4.5GB), 
since it has the largest number 
of feature maps on input.
Nevertheless, checkpointing 
the whole DenseNet unit 
(cat 1$\times$1 3$\times$3)
brings further 3GB.
Finally, checkpointing stem, transition-down 
and transition-up blocks 
relieves additional 1.8GB.
This results in more than a five-fold
reduction of memory requirements, 
from 11.3GB to 2.1B.

Experiments with the label (block)
treat each dense block 
as a checkpoint segment.
This requires more memory than 
(cat 1$\times$1 3$\times$3)
because additional memory 
needs to be allocated
during recomputation. 
The approach (cat 1$\times$1) (3$\times$3) 
is similar to the related previous work \cite{efficientdensenet17arxiv}.
We conclude that the smallest memory footprint
is achieved by checkpointing the stem,
transition-down and transition-up blocks,
as well as DenseNet units as a whole.
Table \ref{tab:segment_params} shows 
that our checkpointing approach 
allows training the LDN161 model
with a six-fold increase of batch size 
with respect to the baseline implementation.
On the other hand, previous 
checkpointing techniques \cite{RotPorKon18a}
yield only a two-fold increase of batch size.

\begin{table}[htb]
\renewcommand{\arraystretch}{1.3}
\caption{Comparison of memory footprint 
  and training speed across various 
  model variants.
  We process 768$\times$768 images
  on a Titan Xp with 12 GB RAM.
  The column 3 (Memory) assumes batch size 6.
}
\label{tab:segment_params}
\footnotesize
\centering
\begin{tabular}{l|c|c|c|c}
  & Uses & Memory & Max & Train \\
  Variant & Ckpt & (MB) & BS & FPS \\
  \hline\hline
  LDN161 32\upmark4 & & 20032 & 3 & 5.6 \\
  ResNet101 32\upmark4 & & 15002 & 4 & 7.8 \\
  LDN121 32\upmark4 & & 11265 & 6 & 11.3 \\
  ResNet50 32\upmark4 & & 10070 & 6 & 11.6 \\
  ResNet18 32\upmark4 & & 3949 & 17 & 24.4 \\
  LDN161 32\upmark4 & \checkmark & 3241 & 19 & 4.4 \\
  LDN121 32\upmark4 & \checkmark  & 2106 & 27 & 9.2 \\
\end{tabular}
\end{table}

\subsection{Qualitative Cityscapes results}
\label{ss:cityq}

\begin{figure}[h]
  \begin{center}
  \includegraphics[width=\columnwidth]{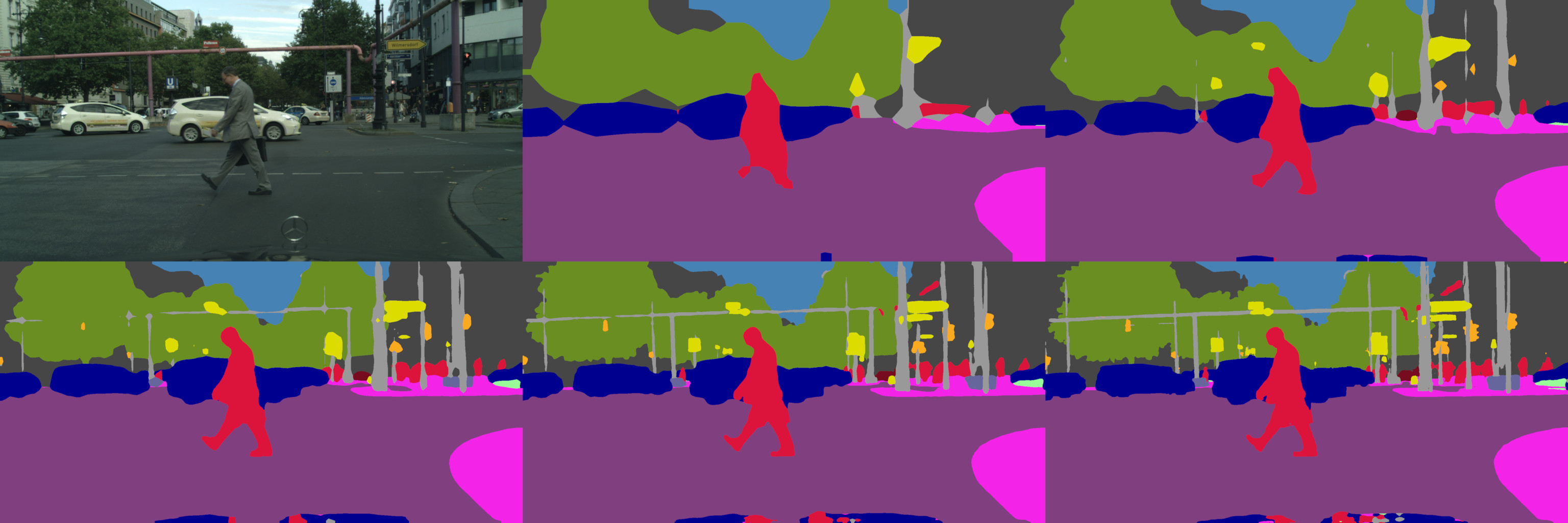}
  
  \vspace{0.5mm}
  \includegraphics[width=\columnwidth]{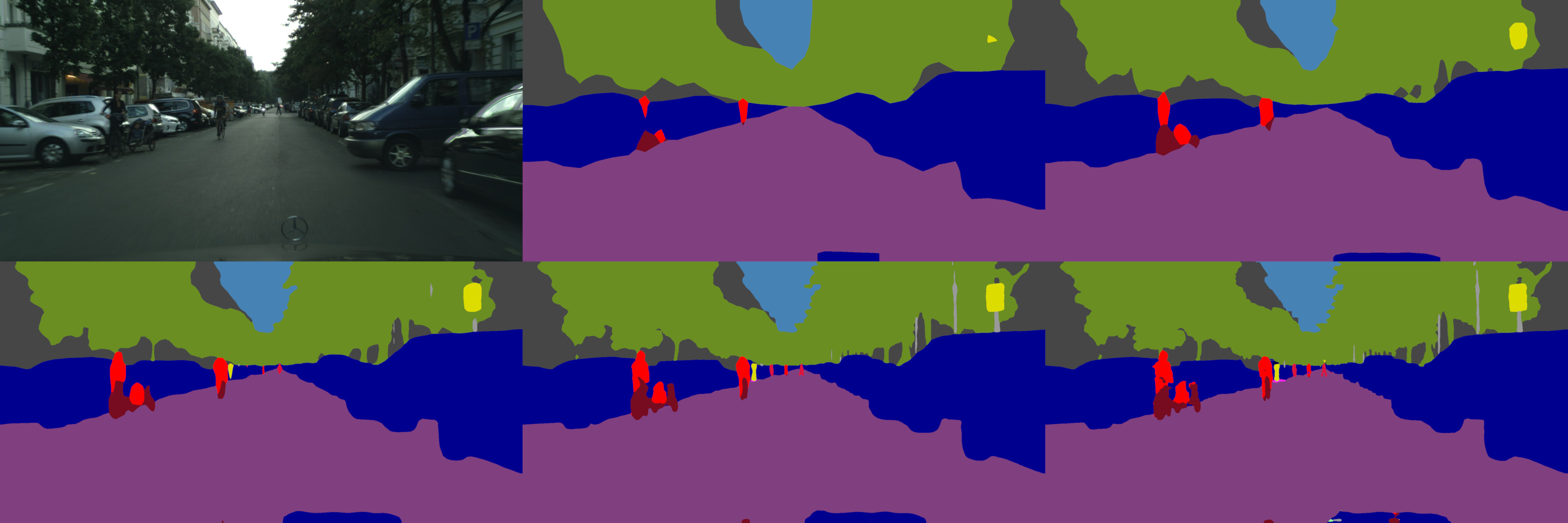}
  \end{center}
  \caption{Impact of ladder-style upsampling
  to the precision at semantic borders 
  and detection of small objects.
  For each of the two input images 
  we show predictions 
  at the following 
  levels of subsampling
  along the upsampling path:
  64$\times$, 32$\times$, 16$\times$, 8$\times$, and 4$\times$.}
  \label{fig:up_path}
\end{figure}

Finally, we present some 
qualitative results on Cityscapes.
Figure \ref{fig:up_path} shows
predictions obtained 
from different layers
along the upsampling path.
Predictions from early layers
miss most small objects,
however ladder-style upsampling
succeeds to recover them due to 
blending with high-resolution features.

\begin{figure}[h]
  \begin{center}
  \includegraphics[width=0.33\columnwidth]{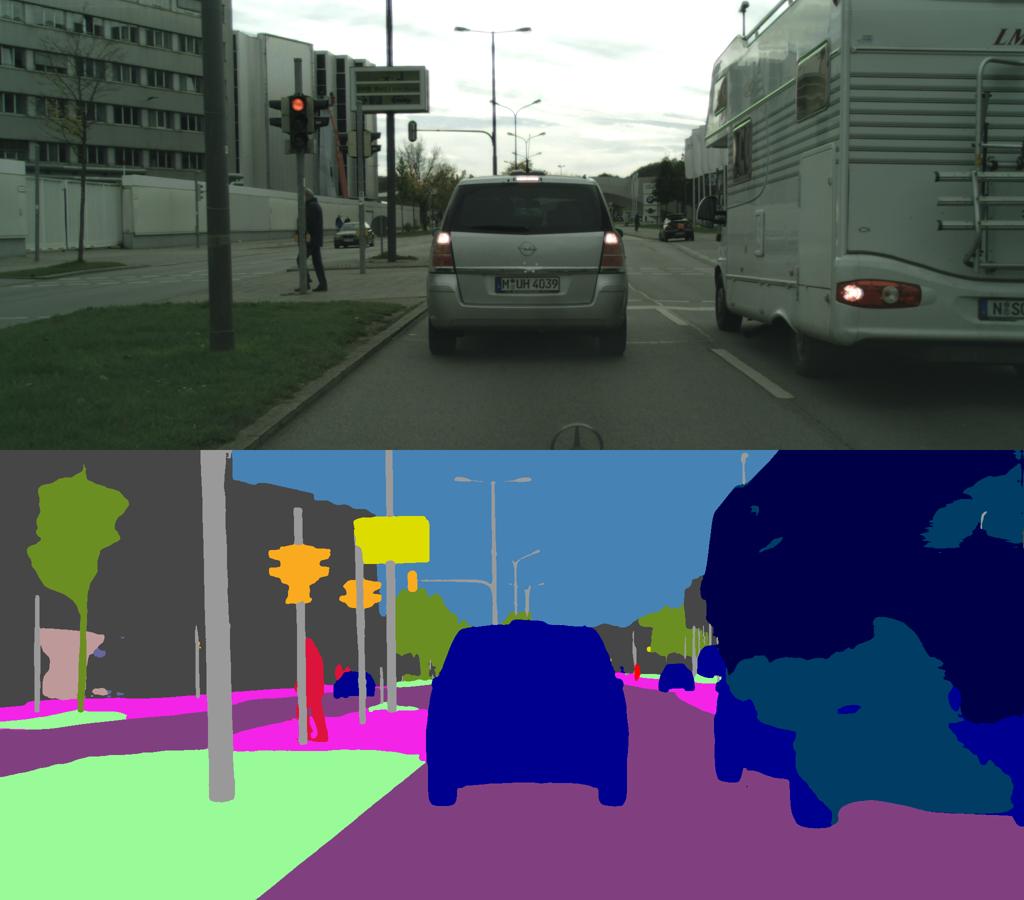}
  \hspace{-1.5mm}
  \includegraphics[width=0.33\columnwidth]{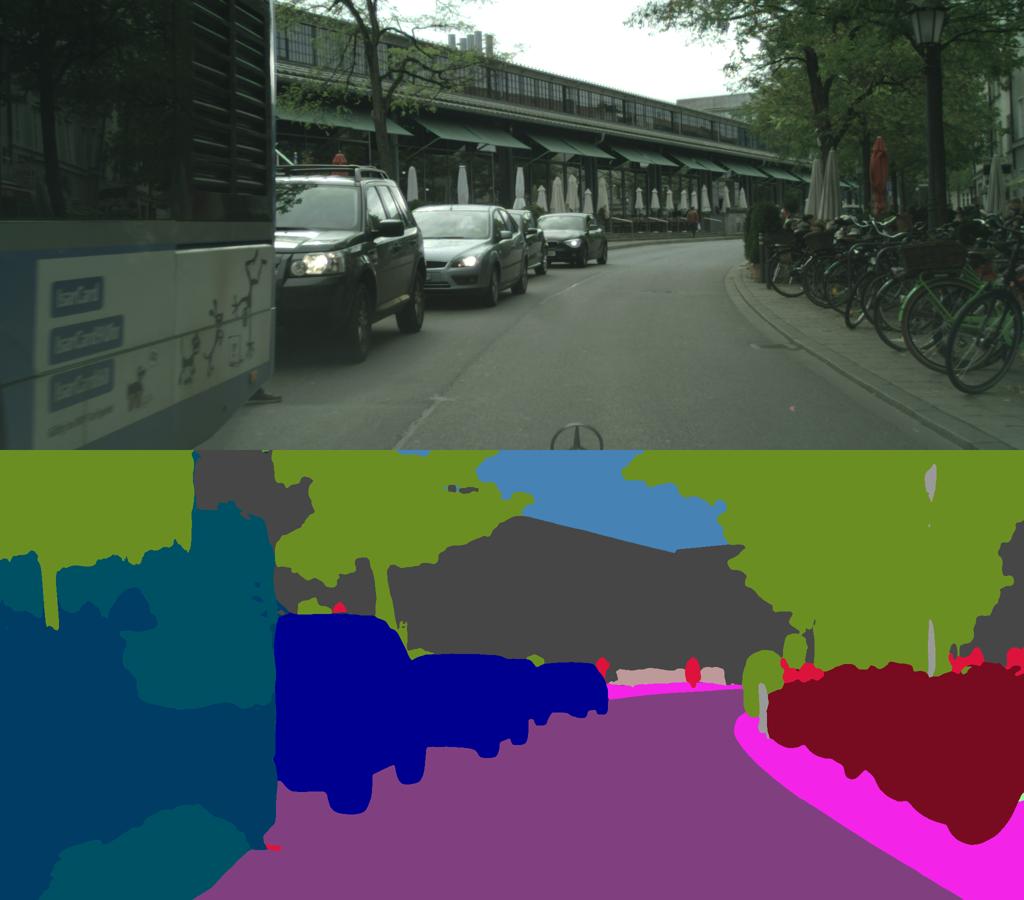}
  \hspace{-1.5mm}
  \includegraphics[width=0.33\columnwidth]{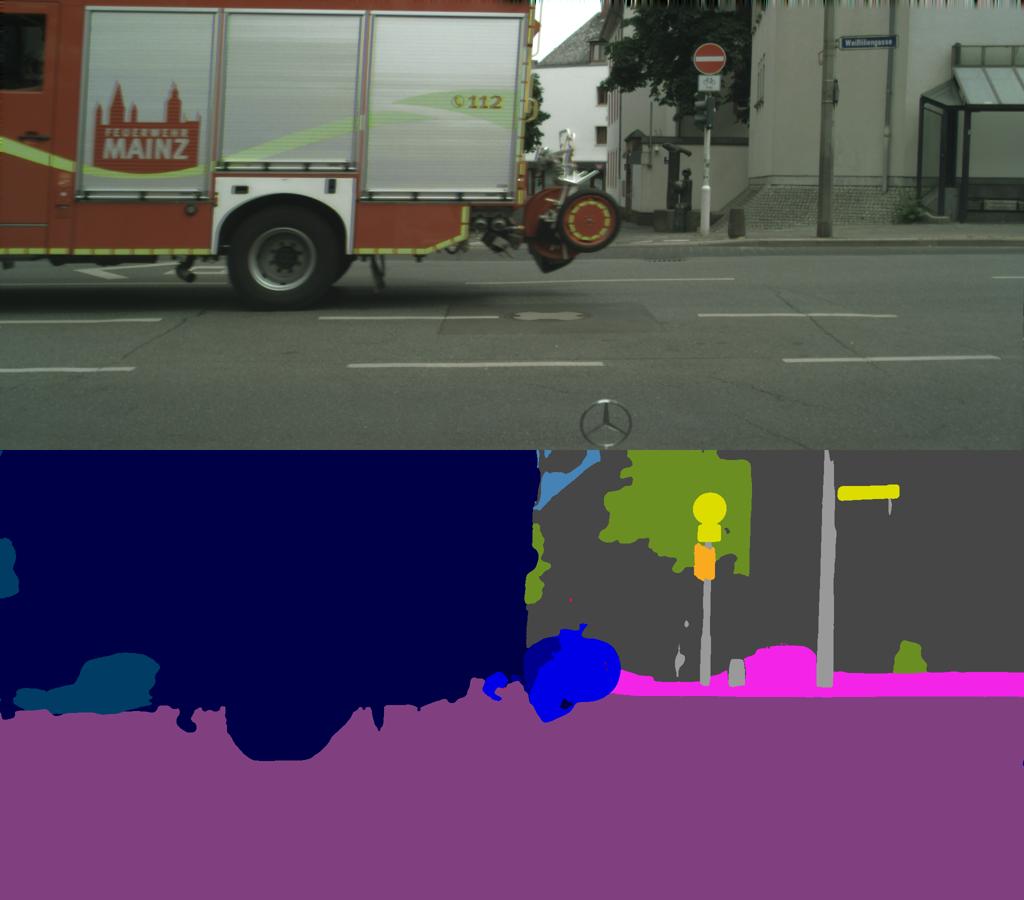}
  
  \vspace{0.5mm}
  \includegraphics[width=0.33\columnwidth]{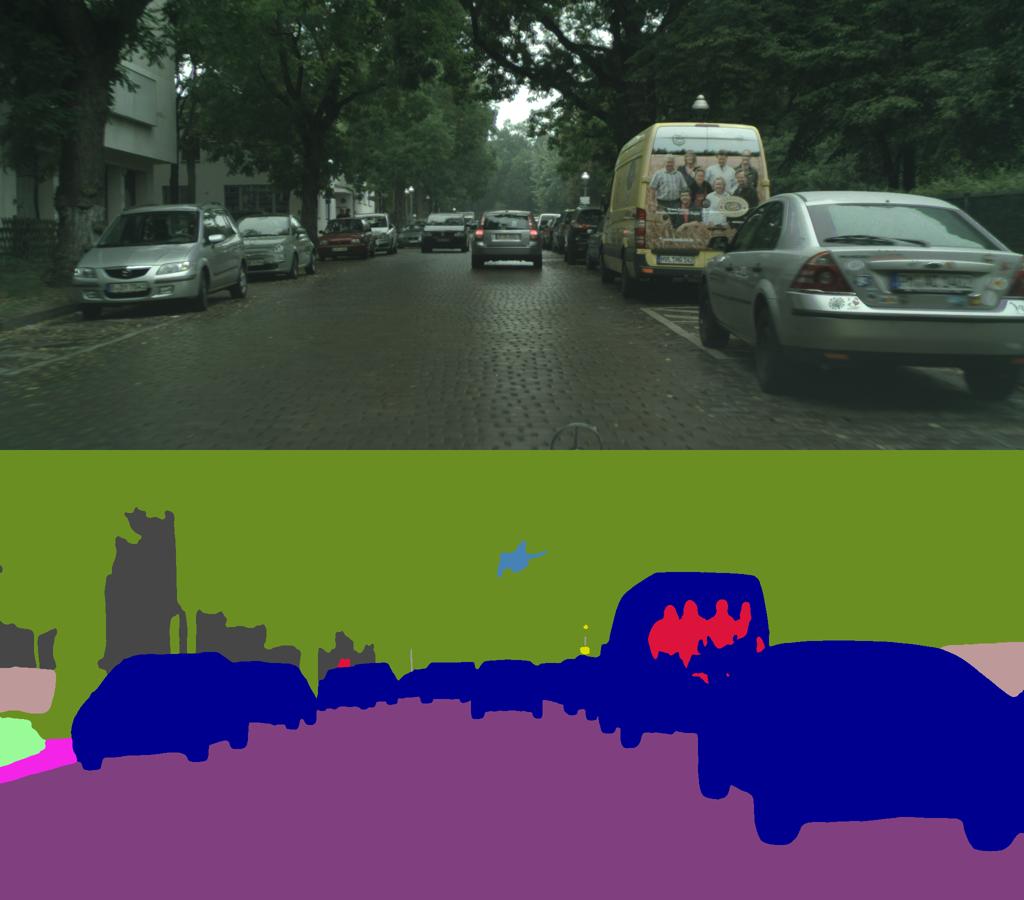}
  \hspace{-1.5mm}
  \includegraphics[width=0.33\columnwidth]{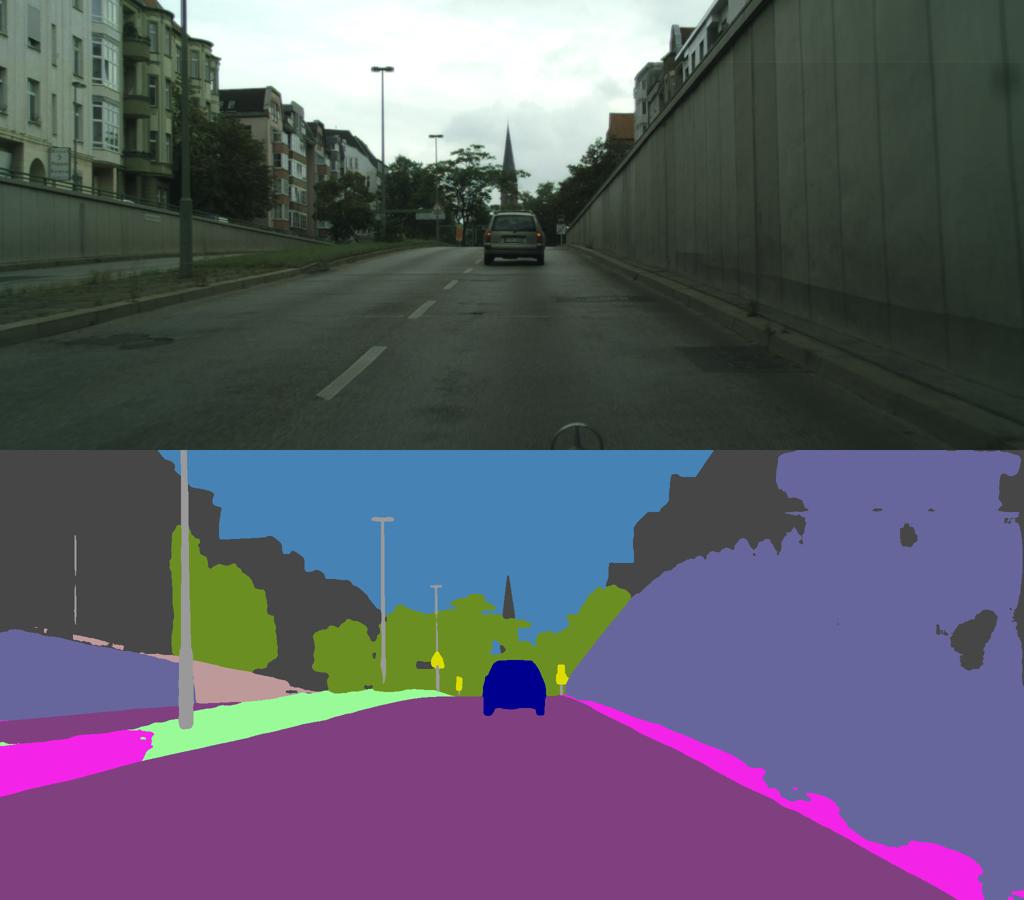}
  \hspace{-1.5mm}
  \includegraphics[width=0.33\columnwidth]{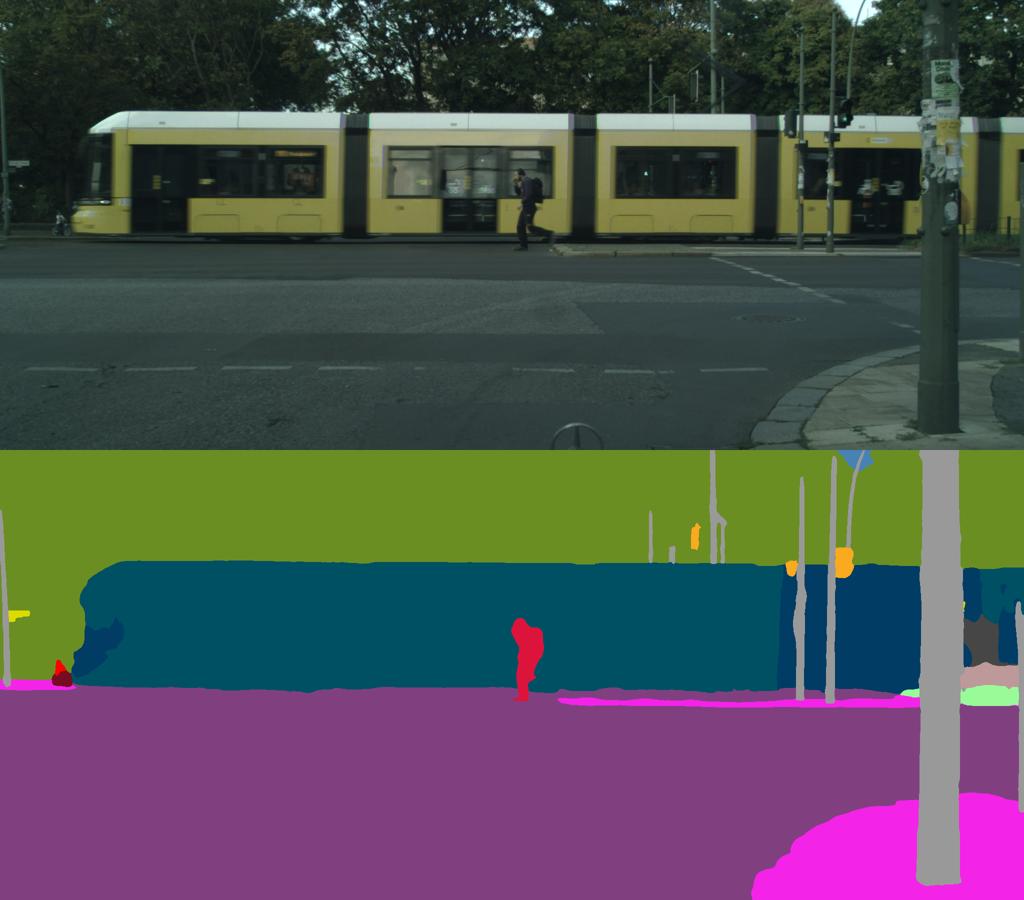}
  \end{center}
  \caption{Images 
    from Cityscapes test 
    where our model misclassified
    some parts of the scene. 
    These predictions are produced by 
    the LDN161 64\upmark4 model
    which achieves 
    80.6\% \mean{IoU} on the test set.
  }
  \label{fig:city_hard}
\end{figure}

Figure \ref{fig:city_hard} shows images 
from Cityscapes test in which
our best model commits the largest mistakes.
It is our impression that 
most of these errors are
due to insufficient context,
despite our efforts to 
enlarge the receptive field.
Overall, we achieve the worst IoU
on fences (60\%) and walls (61\%).

\begin{figure}[h]
  \begin{center}
  \includegraphics[width=0.33\columnwidth]{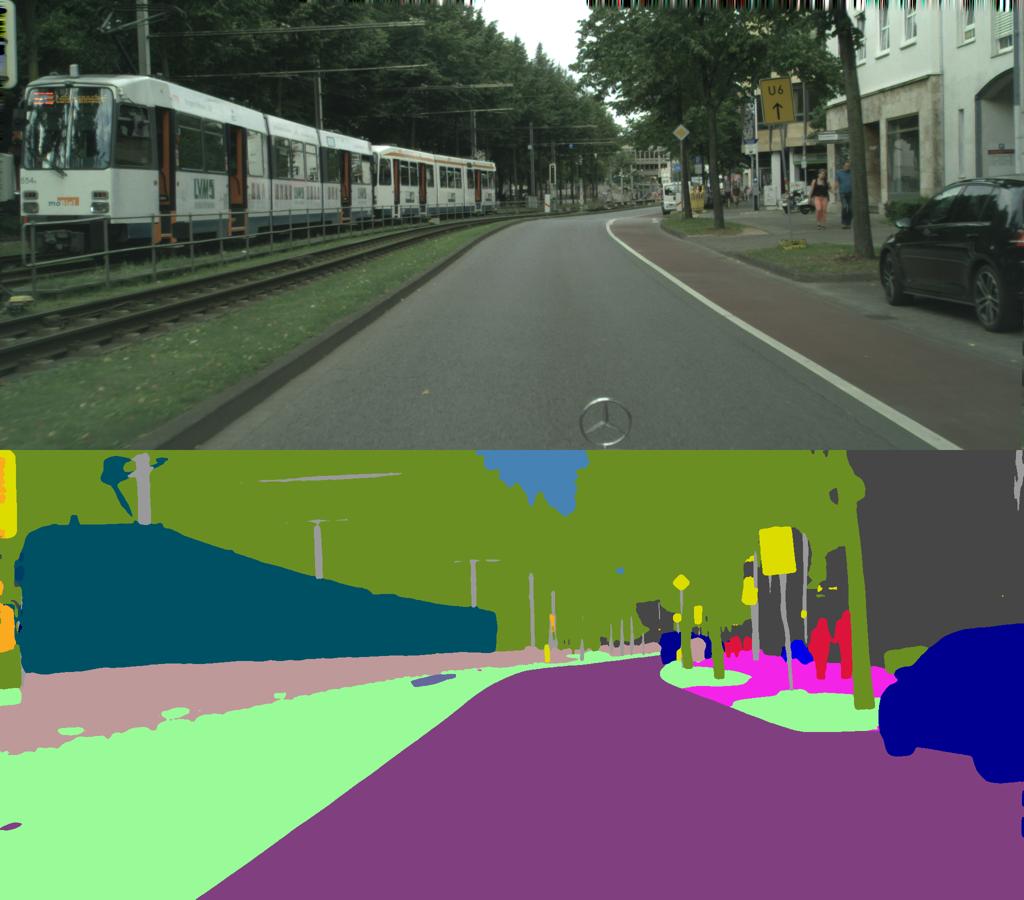}
  \hspace{-1.5mm}
  \includegraphics[width=0.33\columnwidth]{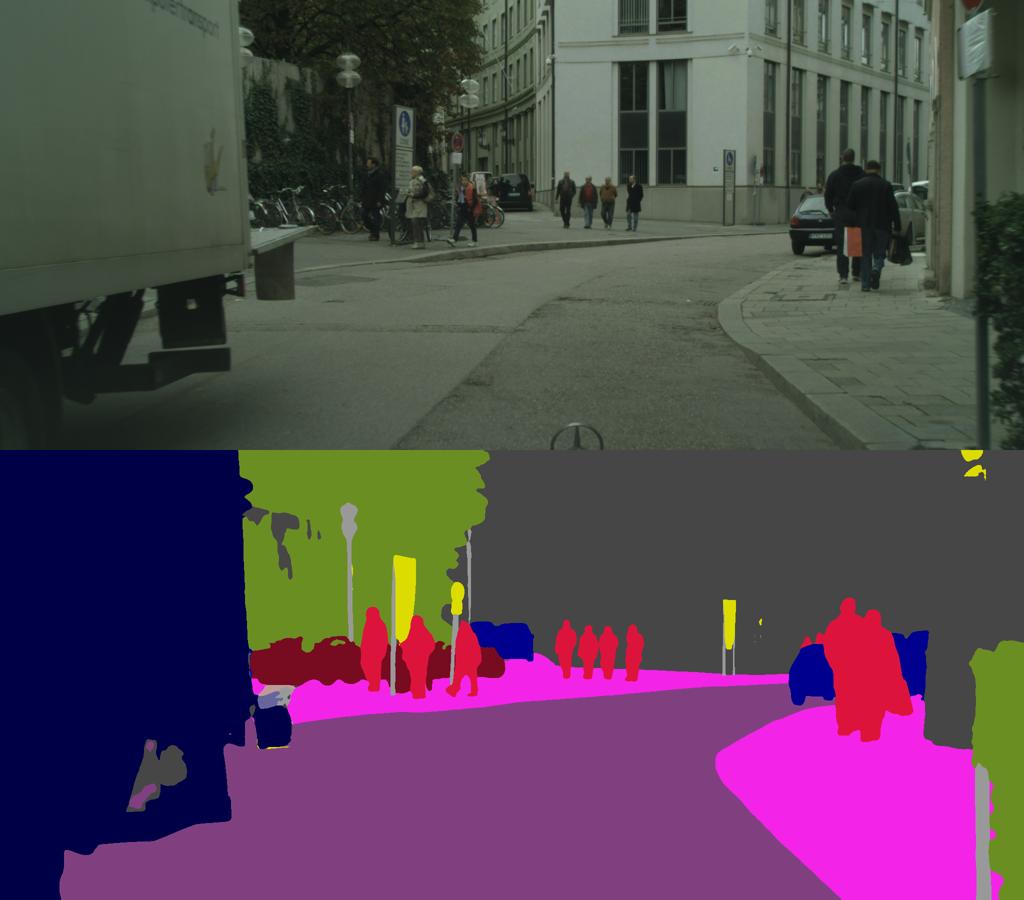}
  \hspace{-1.5mm}
  \includegraphics[width=0.33\columnwidth]{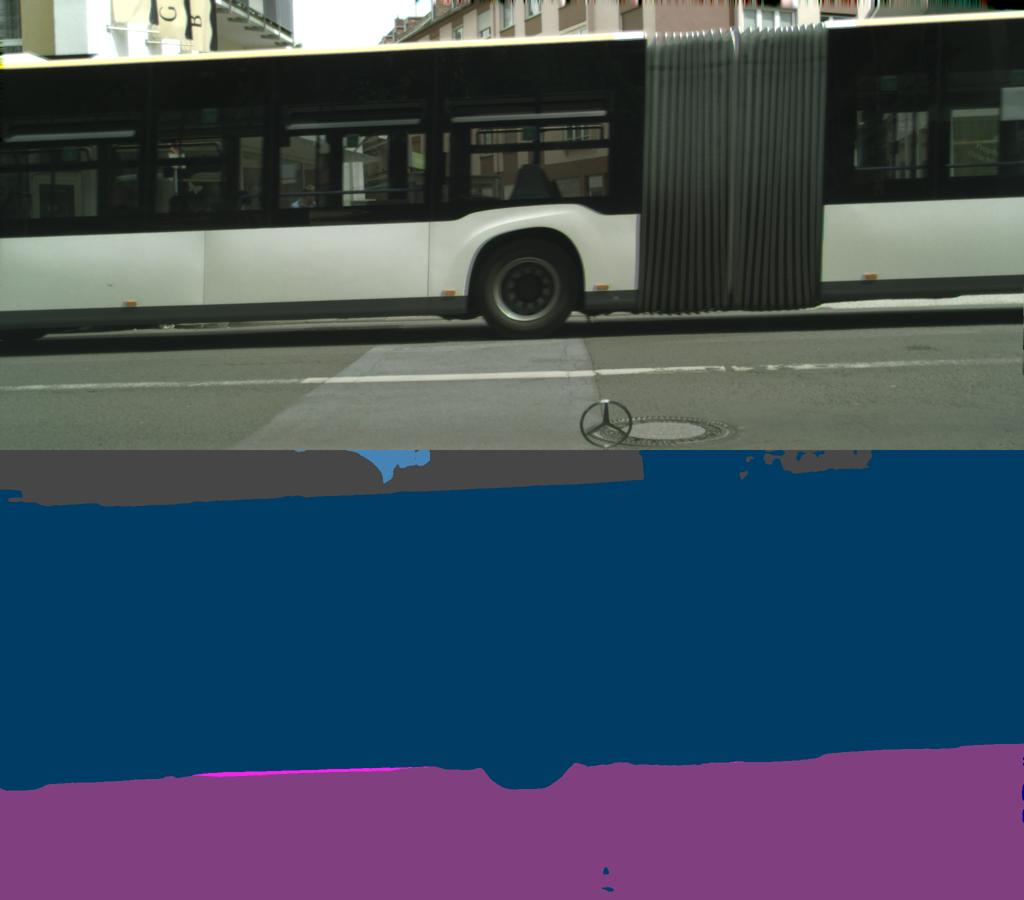}
  
  \vspace{0.5mm}
  \includegraphics[width=0.33\columnwidth]{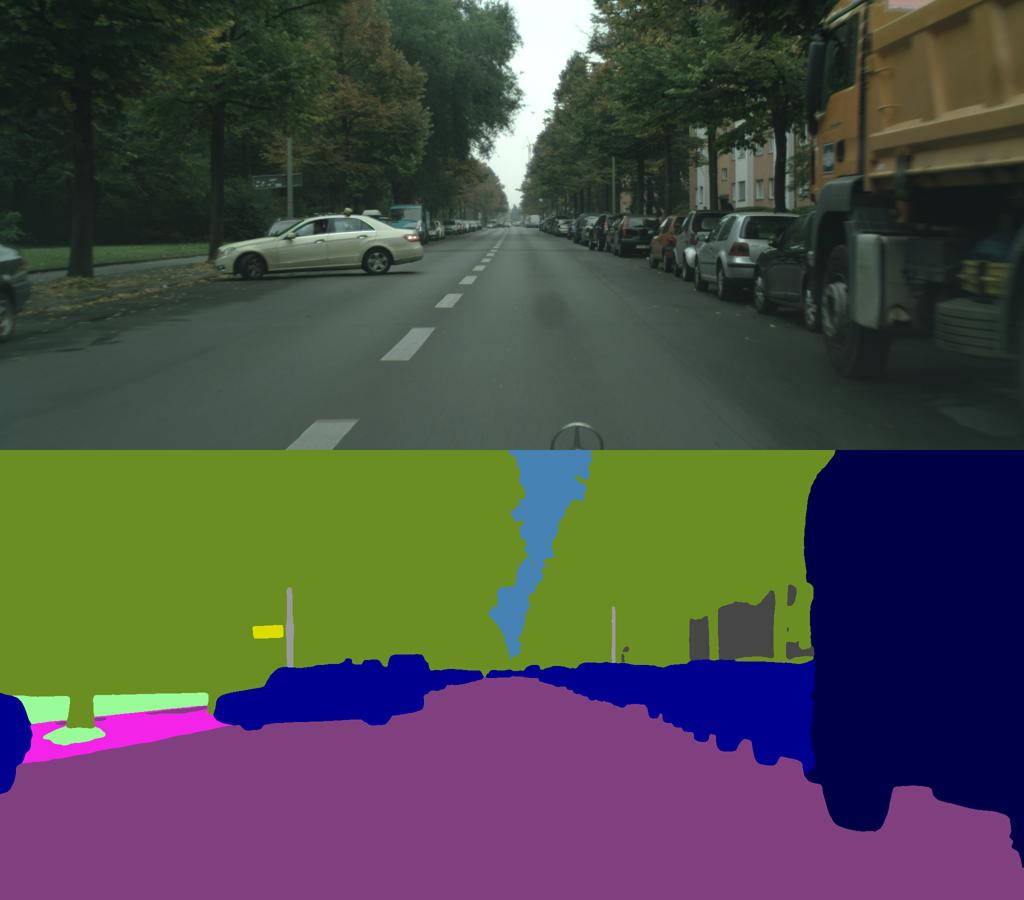}
  \hspace{-1.5mm}
  \includegraphics[width=0.33\columnwidth]{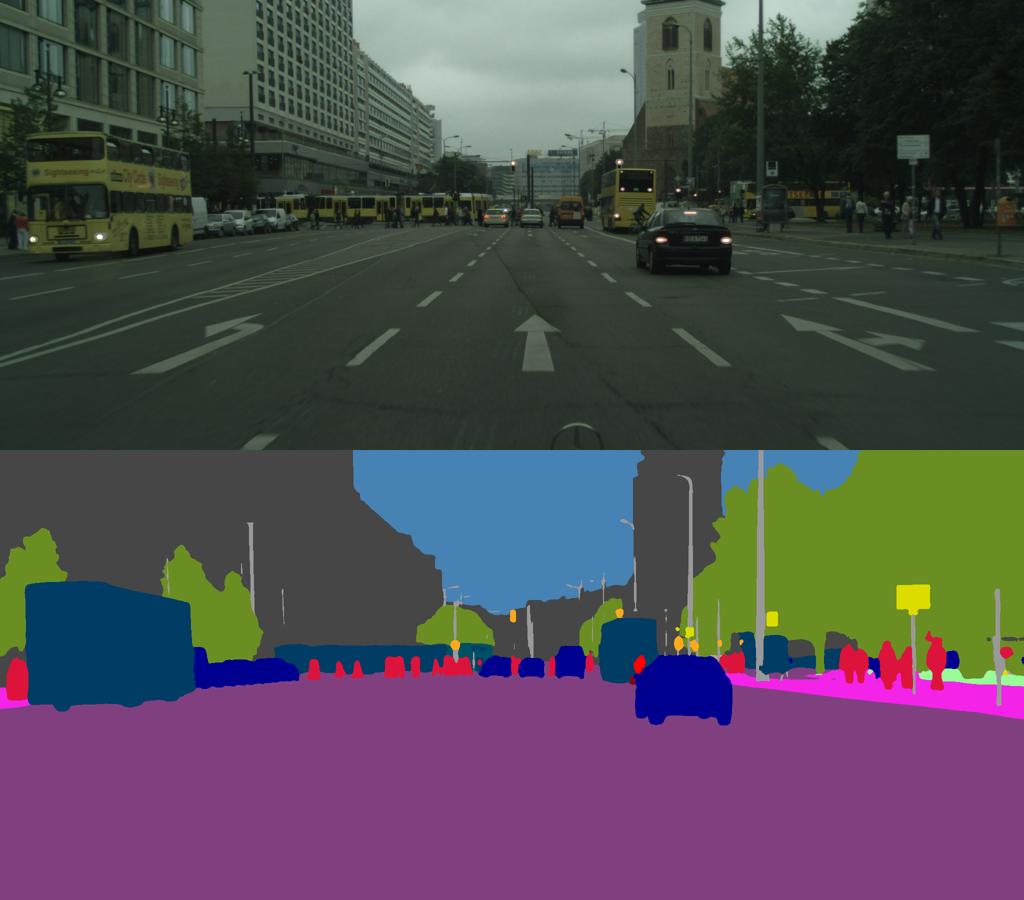}
  \hspace{-1.5mm}
  \includegraphics[width=0.33\columnwidth]{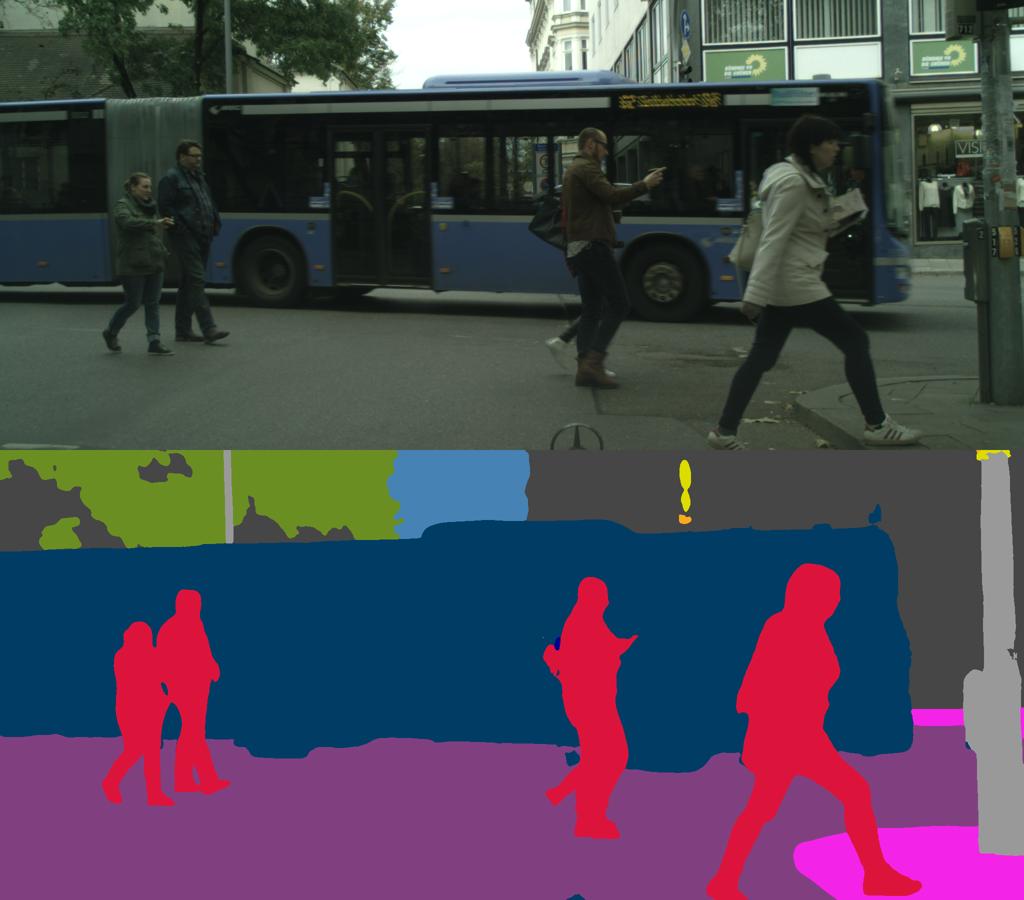}
  \end{center}
  \caption{Images from Cityscapes test
    where our best model 
    (LDN161 64\upmark4, 80.6\% \mean{IoU} test)
    makes no significant errors
    in spite of large objects, 
    occlusions, small objects
    and difficult classes.
  }
  \label{fig:city_good}
\end{figure}

Finally,  Figure \ref{fig:city_good} 
shows some images from Cityscapes test 
where our best model performs well
in spite of occlusions and large objects.
We note that small objects
are very well recognized
which confirms the merit
of ladder-style upsampling.

\section{Conclusion}

We have presented a ladder-style 
adaptation of the DenseNet architecture
for accurate and fast semantic segmentation 
of large images.
The proposed design uses lateral skip connections
to blend the semantics of deep features
with the location accuracy of the early layers.
These connections relieve the deep features
from the necessity to preserve low-level details
and small objects, and allow them to focus 
on abstract invariant and context features.
In comparison with various dilated architectures,
our design substantially decreases 
the memory requirements 
while achieving more accurate results.

We have further reduced the memory requirements
by aggressive gradient checkpointing 
where all batchnorm and projection layers
are recomputed during backprop.
This approach decreases
the memory requirements 
for more than 5 times
while increasing the training time
for only around 20\%.
Consequently, we have been able to train 
on 768$\times$768 crops with batch size 16
while requiring only 5.3 GB RAM.

We achieve state-of-the-art results
on Cityscapes test 
with only fine annotations 
(LDN161: 80.6\% \mean{IoU})
as well as on Pascal VOC 2012 test
without COCO pretraining
(LDN161: 83.6\% \mean{IoU}).
The model based on DenseNet-121
achieves 80.6\% \mean{IoU} on Cityscapes test
while being able 
to perform the forward-pass 
on 512$\times$1024 images in real-time
(59 Hz) on a single Titan Xp GPU.
To the best of our knowledge, 
this is the first account
of applying a DenseNet-based architecture 
for real-time dense prediction 
at Cityscapes resolution.

We have performed extensive experiments 
on Cityscapes, CamVid, ROB 2018 
and Pascal VOC 2012 datasets.
In all cases our best results 
have exceeded or matched  
the state-of-the-art.
Ablation experiments confirm
the utility of the DenseNet design,
ladder-style upsampling and 
aggressive checkpointing.
Future work will exploit 
the reclaimed memory resources
for end-to-end training of
dense prediction in video.
\ifCLASSOPTIONcaptionsoff
  \newpage
\fi



\bibliographystyle{IEEEtran}
\bibliography{IEEEabrv,ms}
\end{document}